# Robust Visual Knowledge Transfer via EDA

Lei Zhang, *Member, IEEE* and David Zhang, *Fellow, IEEE*

*Abstract*—We address the problem of visual knowledge adaptation by leveraging labeled patterns from source domain and a very limited number of labeled instances in target domain to learn a robust classifier for visual categorization. This paper proposes a new extreme learning machine based cross-domain network learning framework, that is called Extreme Learning Machine (ELM) based Domain Adaptation (EDA). It allows us to learn a category transformation and an ELM classifier with random projection by minimizing the $\ell_{2,1}$-norm of the network output weights and the learning error simultaneously. The unlabeled target data, as useful knowledge, is also integrated as a fidelity term to guarantee the stability during cross domain learning. It minimizes the matching error between the learned classifier and a base classifier, such that many existing classifiers can be readily incorporated as base classifiers. The network output weights cannot only be analytically determined, but also transferrable. Additionally, a manifold regularization with Laplacian graph is incorporated, such that it is beneficial to semi-supervised learning. Extensively, we also propose a model of multiple views, referred as MvEDA. Experiments on benchmark visual datasets for video event recognition and object recognition, demonstrate that our EDA methods outperform existing cross-domain learning methods.

## I. INTRODUCTION

IN recent years, the computer vision community has witnessed a significant progress in content based video/image retrieval from a large amount of web video and image data. Visual event recognition and object recognition, however, still remain extremely challenging in real-world cross-domain scenarios containing a considerable camera motion, occlusion, cluttered background, geometric and photometric variations, and large intra-class variations within the same category of videos or images [1]-[3]. It violates the basic assumption of machine learning that the test data lies in the same feature space as training data. Additionally, annotating a large number of videos and images also imposes a great challenge to conventional visual recognition tasks.

To address the above issues in video event recognition, Chang *et al.* [4] developed a multimodal consumer video recognition based on visual and audio features obtained from the consumer video dataset [5], in which 25 concepts were manually labeled. Xu *et al.* [6] studied the problem of unconstrained news video event recognition, and proposed a discriminative kernel method with multilevel temporal alignment. Laptev *et al.* [7] investigated the movie scripts for automatic video annotation of human actions, and proposed a multi-channel non-linear support vector machine (SVM) method. Liu *et al.* [8] studied realistic action recognition based on motion and static features from videos in the wild, such that the influence caused by camera motion, changes in object appearance, and scale, etc. is alleviated. In object recognition, Gehler *et al.* [9] studied several feature combination methods including average kernel SVM (AKSVM), product kernel SVM (PKSVM), multiple kernel learning (MKL) [10], column generation boosting (CG-Boost) [11], and linear programming boosting (LP-B and LP-β) [12]. Joint sparse representation and dictionary learning have also been studied for robust object recognition in [13]-[16].

These traditional learning methods for video event recognition [1]-[4], [6]-[8] and object recognition [9]-[16] can achieve promising results when sufficient and labeled training data are provided, and also both training and testing data are drawn from the same domain. However, it is time consuming and expensive to annotate a large number of training data in real-world applications. Consequently, sufficient training data that share the same feature space and statistical properties (*e.g.*, mean, intra-class, and inter-class variance) as the testing data cannot be guaranteed. Therefore, it violates the basic assumption that the training and testing data should be with similar feature distribution. Existing work demonstrate that the mismatch of data distribution may be alleviated by domain adaptation, such as the sampling selection bias [17] or covariate shift [18]. The training instances in the source domain are re-weighted by leveraging some data from the target domain.

In this paper, following [19]-[25], we propose a new cross domain learning framework for computer vision tasks, *e.g.*, consumer video events recognition by leveraging a large number of labeled YouTube videos (web data) and object recognition by leveraging the labeled images collected with different experimental conditions (*e.g.*, camera, angle, illumination, etc.). In Fig. 1, two frames of "sport" are given to visualize the domain shift/bias.

By reviewing the existing works in domain adaptation, the motivations behind the proposed idea are as follows.

This work was supported by National Natural Science Foundation of China under Grant 61401048 and Research fund of Central Universities.

- L. Zhang is with the College of Communication Engineering, Chongqing University, Chongqing 400044, China. (e-mail: leizhang@cqu.edu.cn).
- D. Zhang is with the Department of Computing, Hong Kong Polytechnic University, Hung Hom, Kowloon, Hong Kong. (e-mail: csdzhang@comp.polyu.edu.hk).

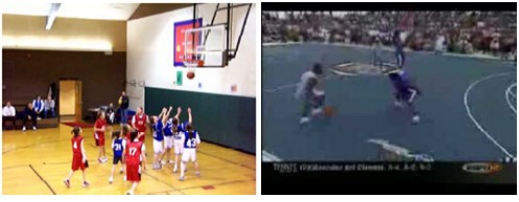

Fig. 1. Two frames from consumer videos (left) and YouTube videos (right).

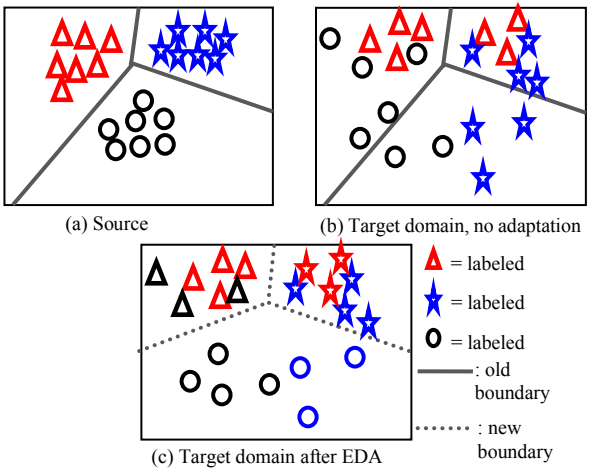

(a) Source      (b) Target domain, no adaptation

(c) Target domain after EDA

△ = labeled
★ = labeled
○ = labeled
—— : old boundary
······ : new boundary

Fig. 2. The data distribution and decision boundaries. (a) linear classifiers learned for a three-class problem on labeled data in source domain. (b) classifiers learned on the source domain do not fit the target domain due to the change of data distribution. (c) domain adaptation with EDA by simultaneously learning new classifier and category transformation matrix. Note that the category transformation denotes the output adaptation with a matrix **Θ**.

- Due to that the distribution difference between source data $\mathbf{X}_S$ and target data $\mathbf{X}_T$ is uncertain, the features from source and target domain are expected to be randomized (randomly corrupted by $\mathbf{W}$, *i.e.*, $\mathbf{WX}_S$ and $\mathbf{WX}_T$) in classifier learning. Then the domain bias of the randomly projected features between two domains can be easily adapted by adjusting the cross-domain classifier with random components.

- It would be interesting to consider an output adaptation in *label* space, not only the conventional input adaptation in feature space. Adaptation in label space can also contribute to the classification performance during the learning process from input to output. This is motivated by the fact that the label space between source and target domains may also be different, *i.e.*, $P(Y_S|\mathbf{X}_S) \neq P(Y_T|\mathbf{X}_T)$. The conventional domain adaptation problem is summarized as *Problem 1*.

- Classifier-based domain adaptation is often with low robustness, because of the uncertain domain variances. It is therefore rational to design a more complex classifier by fusing multiple terms such as cost function, regularizer, etc. in the objective function. However, complex model model result in a high computation complexity. To guarantee the optimization efficiency, an extreme learning machine mechanism with closed-form solution is particularly desired.

With these motivations, the proposed idea is preliminarily described in Fig.2. Fig.2(a) and (b) show the same decision boundary for three classes. Fig.2(b) denotes the inseparability caused by domain shift and Fig.2(c) denotes the newly learned decision boundary via the proposed idea with a well-learned

> **Problem 1 (Domain Adaptation).** *Given a source data, and learn a classifier that can fit a target dataset with different feature distribution and statistical properties (e.g. mean, intra- and inter-class variance), i.e.* $P(\mathbf{X}_S|Y) \neq P(\mathbf{X}_T|Y)$.

*category transformation.* Fig.2(c) implies that our goal is to learn a robust classifier with automatic category transformation learning under "random" domain shift (manually corrupted).

The flowchart of the proposed EDA framework is described in Fig.3 and the merits of this paper are as follows.

- Many existing classifiers including ELM [25]-[27], SVM, MKL [28]-[30], and domain adaptation methods [19]-[22], etc. can be incorporated into the proposed EDA framework as base classifiers accounting for the unlabeled target data, such that the EDA is more flexible.

- Feature transformation is commonly used to reduce the mismatch between data distribution of different domains [23]. However, the probabilistic distribution $P(Y_S)$ and $P(Y_T)$ may also be different. The proposed EDA also concentrates on the consistency of $P(Y)$ for "output/label" adaptation by learning a category transformation, besides only varying the data distribution in feature level.

- To our knowledge, EDA is the first cross-domain learning under the extreme learning machine (ELM) framework, and solved with the basic formulation of ELM. EDA does not assume that the training and testing data are drawn from the same domain. Moreover, by comparing with SVM based domain adaptation, EDA directly learns a classifier for multi-class problem. More importantly, the randomly corrupted features with augmented domain bias can enhance the robustness of the proposed cross-domain classifier.

- Inspired by semi-supervised learning methods [31], [54], a manifold structure preservation term, *i.e.*, manifold regularization based on a graph Laplacian matrix for label consistency, is incorporated into EDA. The intrinsic geometry information of unlabeled data is exploited.

- Multi-view learning [32], [33] concept in the scenario where multiple observations of an image are available is incorporated in EDA. The complementary manifold structural information among features can be exploited for improving the domain adaptation performance.

The remainder of the paper is organized as follows. We briefly review the related work in Section II. Then we introduce our Extreme Learning Machine based Domain Adaptation (EDA) framework in Section III. In Section IV, the EDA is extended to multiple views for dealing with the scenarios with multi-view representations. The experiments and comparisons with state-of-the-art methods based on several benchmark vision datasets are discussed in Section V. Finally, conclusive remarks are provided in Section VI.

## II. RELATED WORKS

### A. Domain Adaptation

Domain adaptation tackles the problems where the distribution over the features varies across tasks (domains) (*e.g.*, data bias [17] and covariate shift [18]) by leveraging labeled data in a related domain when learning a classifier. It has been exploited in natural language processing [34], [35], and recently computer vision [36]-[37], [59]. In this paper, we are committed to the challenging computer vision issues. In what follows, we present recent domain adaptation approaches.

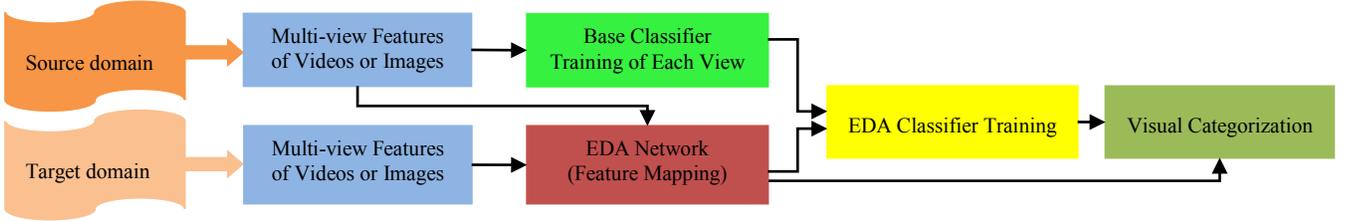

Fig. 3. The flowchart of the proposed cross-domain learning framework in multiple views. Specifically, for each domain, the same type of feature is first extracted. Second, the base classifier is trained on the raw feature of source data. Third, the feature mapping (random projection) is conducted on the both features of source and target data. Fourth, the EDA based domain adaptation classifier is learned. Finally, the visual categorization task with domain adaptation is done.

One of the prominent approaches is feature augmentation, *i.e.,* feature replication (FR) [35], which defines augmented feature vectors in source and target domains. Other feature augmentation based approaches such as semi-supervised heterogeneous feature augmentation (SHFA) [21] and heterogeneous feature augmentation (HFA) [39] learn a feature transformation into a common latent space as well as the classifier. Asymmetric regularized cross-domain transform (ARC-t) method proposed in [36] uses the labeled training data from both domains to learn an asymmetric transformation metric between different feature spaces. Further, in Symm [22] and max-margin domain transforms (MMDT) [22], a category specific feature transformation is learned to diminish the domain bias. Zhang *et al.* [23] propose a latent sparse domain transfer (LSDT) method for sparse subspace reconstruction between source and target domain and achieve state-of-the-art cross-domain performance. Zhang *et al.* [24] also proposed a domain adaptation ELM method for time-varying drift compensation in E-nose. In heterogeneous spectral mapping (HeMap) [41] and domain adaptation manifold alignment (DAMA) [42], a common feature space is learned by utilizing class labels of the source and target training data. Geodesic flow kernel (GFK) [43] aims at purely unsupervised subspace learning in the source and target domain, which shows how to exploit all subspaces on the geodesic path based on kernel trick. In [57], an unsupervised sampling geodesic flow (SGF) is proposed for low-dimensional subspace transfer. The idea behind SGF is that it samples a group of subspaces along the geodesic between source and target domain, and projects the source data into the subspaces. In [55], a LandMark method is proposed for bridging the source and target domain.

In classifier adaptation based cross domain learning, most are based on SVM and MKL. Under the framework of SVM, a transductive SVM (T-SVM) [38], [44] was formulated to learn a classifier using both the labeled data and unlabeled data. Yang *et al.* [45] proposed an adaptive SVM (A-SVM) to learn a new classifier $f^{\mathcal{T}}(\mathbf{x})$ for the target domain by using $f^{\mathcal{T}}(\mathbf{x}) = f^{\mathcal{S}}(\mathbf{x}) + \delta f(\mathbf{x})$, where the classifier $f^{\mathcal{S}}(\mathbf{x})$ is trained with the labeled source samples and $\delta f(\mathbf{x})$ is the perturbation function. Duan *et al.* [46] proposed a domain transfer SVM (DTSVM) which learns a decision function and also attempts to reduce the mismatch between domain distributions measure by maximum mean discrepancy (MMD). Also, two related state-of-the-art cross domain methods based on MKL framework were proposed as adaptive MKL (A-MKL) [20] and domain transfer MKL (DTMKL) [40], which simultaneously learn a SVM classifier and a kernel function by minimizing the distribution mismatch between source and target data.

Transfer learning (TL), known as multi-task learning, is closely related with domain adaptation (DA). TL has been applied in a wide range of vision problems, such as object categorization. TL addresses a slightly different problem that there are multiple output variables $Y_1, \ldots, Y_T$ (*i.e., T* tasks) under a single distribution of the inputs $p(X)$ (*i.e.,* single domain). Comparatively, domain adaptation addresses the learning problem of single task but with multiple domains. DA aims at solving the problem of $p(\mathbf{X}_S|Y) \neq p(\mathbf{X}_T|Y)$ and TL may also solve the problem of $p(Y_S|\mathbf{X}_S) \neq p(Y_T|\mathbf{X}_T)$. From the viewpoint of inclusion relation, DA based methods may be included in TL based methods. In this paper, we target at proposing a novel domain adaptation method.

### B. Extreme Learning Machines

Extreme learning machine (ELM), proposed for training a "generalized" single-layer feed-forward network (SLFN) by analytically determining the output weights $\boldsymbol{\beta}$ between hidden layer and output layer using Moore-Penrose generalized inverse, has been proven to be effective and efficient algorithms for classification and regression [25]-[27]. In contrast to most of the existing approaches, ELMs only update the output weights $\boldsymbol{\beta}$ with randomly generated $L$ hidden layer matrix. The random hidden layer can be produced based on the random input weights $\mathbf{W}$ and biases $\mathbf{B}$, and an activation function $\mathcal{H}(\cdot)$. Specifically, the training of output weights $\boldsymbol{\beta}$ can be transformed into a regularized least square problem solved efficiently and analytically. Briefly, the ELM model is described mathematically as follows.

$$\min_{\boldsymbol{\beta} \in \mathbb{R}^{L \times c}} \frac{1}{2} \|\boldsymbol{\beta}\|^2 + \frac{C}{2} \sum_{i=1}^{N} \|\mathbf{e}_i\|^2$$
$$\text{s.t. } \mathcal{H}(\mathbf{x}_i^{\mathrm{T}} \cdot \mathbf{W} + \mathbf{B}^{\mathrm{T}})\boldsymbol{\beta} = \mathbf{t}_i^{\mathrm{T}} - \mathbf{e}_i^{\mathrm{T}}, i = 1, \ldots, N$$

where $\mathbf{t}_i, \mathbf{e}_i \in \mathbb{R}^c$ denote the label and error vector *w.r.t* the $i$-th training sample, $C$ is a penalty coefficient on the training error, $N$ is the number of training samples, $c$ is the number of classes and $\mathcal{H}(\cdot)$ is the activation function of the hidden layer with $L$ nodes. Note that $\boldsymbol{\beta} \in \mathbb{R}^{L \times c}, \mathbf{x}_i \in \mathbb{R}^d, \mathbf{W} \in \mathbb{R}^{d \times L}$, and $\mathbf{B} \in \mathbb{R}^L$.

From the compact model above, the optimal solution of $\boldsymbol{\beta}$ (output weights) can be analytically determined as follows.

$$\boldsymbol{\beta}^* = \begin{cases} \left(\mathbf{H}^{\mathrm{T}}\mathbf{H} + \frac{\mathbf{I}_L}{C}\right)^{-1} \mathbf{H}^{\mathrm{T}}\mathbf{T}, & N > L \\ \mathbf{H}^{\mathrm{T}}\left(\mathbf{H}\mathbf{H}^{\mathrm{T}} + \frac{\mathbf{I}_L}{C}\right)^{-1} \mathbf{T}, & N < L \end{cases}$$

where $\mathbf{H} = [\mathcal{H}(\mathbf{x}_1^T \cdot \mathbf{W} + \mathbf{B}^T); ...; \mathcal{H}(\mathbf{x}_N^T \cdot \mathbf{W} + \mathbf{B}^T)] \in \mathbb{R}^{N \times L}$ is the hidden layer output, $\mathbf{T} \in \mathbb{R}^{N \times c}$ is the label matrix, $\mathbf{I}_L$ is a $L \times L$ identity matrix. Interested readers can refer to [26] for more details about the deduction of $\boldsymbol{\beta}$.

ELM theories [25], [27], [60], [61] show that hidden neurons need not be adjusted in many applications and the output weights of the networks can be adjusted based on different optimization constraints which are application dependent. Hidden neurons can be randomly generated independent of the training data or can be transferred from other ancestors. ELM may bridge the gap between machine learning and biological learning. Additionally, the biological learning mechanism of ELM has been confirmed in [62]. However, ELMs with different versions are only studied in single domain and lack of cross domain transfer capability.

## III. EXTREME LEARNING MACHINE BASED DOMAIN ADAPTATION FRAMEWORK (EDA)

### A. Summary of Main Notations

Let $\mathbf{X}_\mathcal{S} \in \mathbb{R}^{d \times N_\mathcal{S}}$ be the data matrix of source domain $\mathcal{S}$, $\mathbf{T}_\mathcal{S} = (\mathbf{t}_\mathcal{S}^1, \cdots, \mathbf{t}_\mathcal{S}^{N_\mathcal{S}})^T \in \mathbb{R}^{N_\mathcal{S} \times c}$ be the label matrix of source domain with $c$ categories, $\mathbf{X}_\mathcal{T} \in \mathbb{R}^{d \times (N_{\mathcal{T}\ell} + N_{\mathcal{T}u})}$ be the data matrix of target domain $\mathcal{T}$, $\mathbf{T}_\mathcal{T} = (\mathbf{t}_{\mathcal{T}\ell}^1, \cdots, \mathbf{t}_{\mathcal{T}\ell}^{N_{\mathcal{T}\ell}})^T \in \mathbb{R}^{N_{\mathcal{T}\ell} \times c}$ be the label matrix of labeled target domain with $c$ categories, and $\mathbf{H}_\mathcal{S} \in \mathbb{R}^{N_\mathcal{S} \times L}$, $\mathbf{H}_\mathcal{T} \in \mathbb{R}^{N_{\mathcal{T}\ell} \times L}$, $\mathbf{H}_{\mathcal{T}u} \in \mathbb{R}^{N_{\mathcal{T}u} \times L}$ and $\mathbf{H} \in \mathbb{R}^{(N_{\mathcal{T}\ell} + N_{\mathcal{T}u}) \times L}$ denote the hidden layer output matrix of source data, labeled target data, unlabeled target data, and all target data, respectively, with $L$ hidden nodes. Let $\boldsymbol{\beta} \in \mathbb{R}^{L \times c}$ be the learned classifier, $\boldsymbol{\Theta} \in \mathbb{R}^{c \times c}$ be the learned category transformation matrix (for output adaptation in label space), $\mathbf{y}_\mathcal{S}^i |_{\boldsymbol{\beta}}$ be the predicted label w.r.t. the $i$-th labeled source sample, $\mathbf{y}_{\mathcal{T}\ell}^j |_{\boldsymbol{\beta}}$ be the predicted label w.r.t. the $j$-th labeled target sample. Let $\boldsymbol{\phi}_p$ be a pre-learned classifier (i.e., base classifier trained on the source data), $\boldsymbol{\phi}_{p,\mathcal{T}u} = (\boldsymbol{\phi}_{p,\mathcal{T}u}^1, \cdots, \boldsymbol{\phi}_{p,\mathcal{T}u}^{N_{\mathcal{T}u}})$ be the predicted target label matrix based on the pre-classifier, where existing classifiers can be used (incorporated) to learn the pre-classifier $\boldsymbol{\phi}_p$ such as ELM, SVM, their variants, etc.

### B. Problem Formulation

In the proposed EDA, the classifier and category transformation matrix are simultaneously learned with domain adaptation ability. Intuitively, three parts related with respect to the source data, labeled target data and unlabeled target data are designed in the EDA, which is generally formulated as

$$\min_{\boldsymbol{\beta}, \boldsymbol{\Theta}} \mathcal{R}_\mathcal{S}(\boldsymbol{\beta}) + \mathcal{V}_{\mathcal{T}\ell}(\boldsymbol{\beta}, \boldsymbol{\Theta}) + \mathcal{V}_{\mathcal{T}u}(\boldsymbol{\beta}, \boldsymbol{\phi}_p) \quad (1)$$

The first term $\mathcal{R}_\mathcal{S}(\boldsymbol{\beta})$ carries out the classifier training using labeled samples from source domain, formulated as

$$\mathcal{R}_\mathcal{S}(\boldsymbol{\beta}) = \|\boldsymbol{\beta}\|_{q,p}^p + C_\mathcal{S} \sum_{i=1}^{N_\mathcal{S}} \left\| \boldsymbol{\xi}_\mathcal{S}^i \right\|^2 \quad (2)$$

where $\|\cdot\|_{q,p}^p$ is used to control the complexity of the output weights $\boldsymbol{\beta}$ (i.e., classifier parameters). $C_\mathcal{S}$ is the penalty coefficient and $\boldsymbol{\xi}_\mathcal{S}^i = \mathbf{t}_\mathcal{S}^i - \mathbf{y}_\mathcal{S}^i |_{\boldsymbol{\beta}}$ denotes the error. In this paper, $\mathbf{t}_\mathcal{S}^{i,j} = 1$ if pattern $\mathbf{x}_i$ belongs to the $j$-th class, and -1 otherwise. $\|\cdot\|_{q,p}$ denotes $\ell_{q,p}$-norm. Given a matrix $\mathbf{Q} \in \mathbb{R}^{m \times n}$, there is

$$\|\mathbf{Q}\|_{q,p} = \left( \sum_{i=1}^m \left( \sum_{j=1}^n |Q_{ij}|^q \right)^{p/q} \right)^{1/p} \quad (3)$$

As can be seen from the Eq.(3), it is common Frobenius norm or $\ell_2$-norm when $p = q = 2$, and in this case the first term of Eq.(2) becomes the conventional ELM. In order to impose sparse property on $\boldsymbol{\beta}$, we constrain $q \geq 2$ and $0 \leq p \leq 2$. Intrinsically, different selection of $(q, p)$ denotes different approaches. If $p = 0$, the formulated problem is not convex and hard to solve, therefore, we suppose $p = 1$. Since $q$ is set to measure the norm of each row vector, $q \geq 2$ is generally used. In this paper, $q = 2$ is set because larger $q$ value does not improve the final results [47]. Therefore, $\ell_{2,1}$-norm of the output weights $\boldsymbol{\beta}$, i.e., $\|\boldsymbol{\beta}\|_{2,1}$ is used in the proposed EDA framework for better sparsity and generalization ability.

The second term $\mathcal{V}_{\mathcal{T}\ell}(\boldsymbol{\beta}, \boldsymbol{\Theta})$ tends to learn the category transformation matrix $\boldsymbol{\Theta}$ and the cross-domain classifier $\boldsymbol{\beta}$ using a few number of labeled target data. It is formulated as

$$\mathcal{V}_{\mathcal{T}\ell}(\boldsymbol{\beta}, \boldsymbol{\Theta}) = C_\mathcal{T} \sum_{j=1}^{N_\mathcal{T}} \left\| \boldsymbol{\xi}_\mathcal{T}^j |_{\boldsymbol{\beta}, \boldsymbol{\Theta}} \right\|^2 + \gamma \|\boldsymbol{\Theta} - \mathbf{I}\|_F^2 \quad (4)$$

where $\boldsymbol{\xi}_\mathcal{T}^j |_{\boldsymbol{\beta}, \boldsymbol{\Theta}} = (\mathbf{t}_\mathcal{T}^j)^T \circ \boldsymbol{\Theta} - \mathbf{y}_{\mathcal{T}\ell}^j |_{\boldsymbol{\beta}}$, and $\|\boldsymbol{\Theta} - \mathbf{I}\|_F^2$ is to control the category distortion during transformation. The symbol $\circ$ denotes a multiplication operator of category transformation via $\boldsymbol{\Theta}$, which is different from feature transformation (input adaptation) for feature alignment between domains [23]. $C_\mathcal{T}$ and $\gamma$ are trade-off parameters. Actually, the category transformation performs output adaptation and makes the classification more conducive. That is, the discrepancy of label distribution between domains can also be aligned in this work in addition to aligning the feature distribution.

In most domain adaptation methods, numerous available unlabeled data in target domain that also have significant contribution to classifier learning are not fully exploited. The importance of unlabeled data has been emphasized in [48]. In this work, for exploiting the unlabeled data, we introduce a fidelity term $\mathcal{V}_{\mathcal{T}u}(\boldsymbol{\beta}, \boldsymbol{\phi}_p)$ to guarantee the generalization and stability of EDA by minimizing the systematic perturbation error between the extreme classifier $\boldsymbol{\beta}$ and the pre-learned classifier $\boldsymbol{\phi}_p$ (e.g., SVM, nearest neighbors, etc.) when fed into the same inputs, which is formulated as

$$\mathcal{V}_{\mathcal{T}u}(\boldsymbol{\beta}, \boldsymbol{\phi}_p) = \tau \sum_{k=1}^{N_{\mathcal{T}u}} \left\| \boldsymbol{\xi}_{\mathcal{T}u}^k |_{\boldsymbol{\beta}, \boldsymbol{\phi}_p} \right\|^2 + \Omega_\mathcal{M}(\boldsymbol{\beta}) \quad (5)$$

where $\boldsymbol{\xi}_{\mathcal{T}u}^k |_{\boldsymbol{\beta}, \boldsymbol{\phi}_p} = \boldsymbol{\phi}_{p,\mathcal{T}u}^k - \mathbf{y}_{\mathcal{T}u}^k |_{\boldsymbol{\beta}}$, and $\Omega_\mathcal{M}(\boldsymbol{\beta})$ denotes the manifold regularization, which as commented in [54], is incorporated in our EDA to improve the classifier adaptability.

To better represent $\Omega_\mathcal{M}(\boldsymbol{\beta})$, we assume that two points $\mathbf{x}_i$ and $\mathbf{x}_j$ are close to each other, then the conditional probability $P(\mathbf{y}|\mathbf{x}_i)$ and $P(\mathbf{y}|\mathbf{x}_j)$ should be similar, which is a widely known smoothness assumption in machine learning. The manifold regularization framework proposed to enforce such assumption, is formulated as

$$\min \frac{1}{2} \sum_{i,j} \mathcal{A}_{i,j} \left\| P(\mathbf{y}|\mathbf{x}_i) - P(\mathbf{y}|\mathbf{x}_j) \right\|^2 \quad (6)$$

where $w_{i,j}$ is the pair-wise similarity between pattern $\mathbf{x}_i$ and $\mathbf{x}_j$. The similarity matrix $\mathcal{A}$ is sparse in which a non-zero element (e.g., 1) is assigned if $\mathbf{x}_i$ is among the $k$ nearest neighbors of $\mathbf{x}_j$, i.e., $\mathbf{x}_i \in \mathcal{N}_k(\mathbf{x}_j)$, or $\mathbf{x}_j$ is among the $k$ nearest neighbors of $\mathbf{x}_i$, i.e., $\mathbf{x}_j \in \mathcal{N}_k(\mathbf{x}_i)$. Due to the difficulty in computing $P(\mathbf{y}|\mathbf{x}_i)$, the Eq.(6) is generally transformed as

$$\min \frac{1}{2}\sum_{i,j}\mathcal{A}_{i,j}\left\|\mathbf{y}_i-\mathbf{y}_j\right\|^2 \tag{7}$$

where $\mathbf{y}_i$ and $\mathbf{y}_j$ are the predicted output label vector *w.r.t.* pattern $\mathbf{x}_i$ and $\mathbf{x}_j$, respectively. By expanding Eq.(7) in matrix trace-form, the manifold structure preservation term $\Omega_{\mathcal{M}}(\boldsymbol{\beta})$ in Eq.(5) is therefore formulated as

$$\Omega_{\mathcal{M}}(\boldsymbol{\beta})=\lambda\cdot tr(\mathbf{F}^{\mathrm{T}}\mathcal{L}\mathbf{F}) \tag{8}$$

where $\mathcal{L}=\mathcal{D}-\mathcal{A}$ is the Laplacian graph matrix, $\mathcal{D}$ is a diagonal matrix with diagonal entries $\mathcal{D}_{ii}=\sum_{j}\mathcal{A}_{i,j}$, and $\mathbf{F}=\left(\mathbf{y}_1,\cdots,\mathbf{y}_{N_{\mathcal{T}\ell}+N_{\mathcal{T}u}}\right)^{\mathrm{T}}$ denotes the target output matrix.

By substituting (2), (4), (5) and (8) into (1), the proposed EDA framework with constraints is summarized as follows

$$\min_{\boldsymbol{\beta},\boldsymbol{\Theta},\boldsymbol{\xi}_{\mathcal{S}}^i,\boldsymbol{\xi}_{\mathcal{T}}^j,\boldsymbol{\xi}_{\mathcal{T}u}^k}\|\boldsymbol{\beta}\|_{2,1}+C_{\mathcal{S}}\sum_{i=1}^{N_{\mathcal{S}}}\left\|\boldsymbol{\xi}_{\mathcal{S}}^i\right\|_2^2+C_{\mathcal{T}}\sum_{j=1}^{N_{\mathcal{T}}}\left\|\boldsymbol{\xi}_{\mathcal{T},\boldsymbol{\Theta}}^j\right\|_2^2+$$
$$\gamma\|\boldsymbol{\Theta}-\mathbf{I}\|_2^2+\tau\sum_{k=1}^{N_{\mathcal{T}u}}\left\|\boldsymbol{\xi}_{\mathcal{T}u}^k|_{\boldsymbol{\beta},\boldsymbol{\Phi}_p}\right\|_2^2+\lambda\cdot tr(\mathbf{F}^{\mathrm{T}}\mathcal{L}\mathbf{F}) \tag{9}$$

$$\text{s.t.}\begin{cases}\mathbf{H}_{\mathcal{S}}^i\boldsymbol{\beta}=\mathbf{t}_{\mathcal{S}}^i-\boldsymbol{\xi}_{\mathcal{S}}^i,i=1,\ldots,N_{\mathcal{S}}\\\mathbf{H}_{\mathcal{T}}^j\boldsymbol{\beta}=\left(\mathbf{t}_{\mathcal{T}}^j\right)^{\mathrm{T}}\circ\boldsymbol{\Theta}-\boldsymbol{\xi}_{\mathcal{T}}^j|_{\boldsymbol{\beta},\boldsymbol{\Theta}},j=1,\ldots,N_{\mathcal{T}\ell}\\\mathbf{H}_{\mathcal{T}u}^k\boldsymbol{\beta}=\boldsymbol{\Phi}_{p,\mathcal{T}u}^k-\boldsymbol{\xi}_{\mathcal{T}u}^k|_{\boldsymbol{\beta},\boldsymbol{\Phi}_p},k=1,\ldots,N_{\mathcal{T}u}\\\mathbf{F}=\mathbf{H}\boldsymbol{\beta}\end{cases} \tag{10}$$

By substituting the constraints (10) into the objective function (9), the EDA model can be compactly rewritten as

$$\min_{\boldsymbol{\beta},\boldsymbol{\Theta}}\mathcal{J}(\boldsymbol{\beta},\boldsymbol{\Theta})=\|\boldsymbol{\beta}\|_{2,1}+C_{\mathcal{S}}\|\mathbf{H}_{\mathcal{S}}\boldsymbol{\beta}-\mathbf{T}_{\mathcal{S}}\|_{\mathrm{F}}^2+C_{\mathcal{T}}\|\mathbf{H}_{\mathcal{T}}\boldsymbol{\beta}-\mathbf{T}_{\mathcal{T}}\circ\boldsymbol{\Theta}\|_{\mathrm{F}}^2+\gamma\|\boldsymbol{\Theta}-\mathbf{I}\|_{\mathrm{F}}^2+\tau\left\|\mathbf{H}_{\mathcal{T}u}\boldsymbol{\beta}-\boldsymbol{\phi}_{p,\mathcal{T}u}\right\|_{\mathrm{F}}^2+\lambda\cdot tr(\boldsymbol{\beta}^{\mathrm{T}}\mathbf{H}^{\mathrm{T}}\mathcal{L}\mathbf{H}\boldsymbol{\beta}) \tag{11}$$

### C. Learning Algorithm

As can be seen from the objective function (11) of EDA, it is differentiable *w.r.t.* $\boldsymbol{\beta}$ and $\boldsymbol{\Theta}$, such that an efficient alternative optimization can be easily proposed to solve this problem.

First, fix $\boldsymbol{\Theta}=\mathbf{I}$, by calculating the derivative of objective function *w.r.t.* $\boldsymbol{\beta}$, we then have

$$\frac{\partial\mathcal{J}(\boldsymbol{\beta},\boldsymbol{\Theta})}{\partial\boldsymbol{\beta}}=2\mathbf{U}\boldsymbol{\beta}+2C_{\mathcal{S}}\mathbf{H}_{\mathcal{S}}^{\mathrm{T}}(\mathbf{H}_{\mathcal{S}}\boldsymbol{\beta}-\mathbf{T}_{\mathcal{S}})+2C_{\mathcal{T}}\mathbf{H}_{\mathcal{T}}^{\mathrm{T}}(\mathbf{H}_{\mathcal{T}}\boldsymbol{\beta}-\mathbf{T}_{\mathcal{T}}\boldsymbol{\Theta})+\tau\mathbf{H}_{\mathcal{T}u}^{\mathrm{T}}(\mathbf{H}_{\mathcal{T}u}\boldsymbol{\beta}-\boldsymbol{\phi}_{p,\mathcal{T}u})+\lambda\mathbf{H}^{\mathrm{T}}\mathcal{L}\mathbf{H}\boldsymbol{\beta} \tag{12}$$

where $\mathbf{U}\in\mathbb{R}^{L\times L}$ is a diagonal matrix, whose $i$-th diagonal element is shown as

$$\mathbf{U}_{ii}=\frac{1}{2\|\boldsymbol{\beta}_i\|_2} \tag{13}$$

where $\boldsymbol{\beta}_i$ denotes the $i$-th row of $\boldsymbol{\beta}$.

In terms of the first term of (12), $\|\boldsymbol{\beta}\|_{2,1}$ can be written as

$$\|\boldsymbol{\beta}\|_{2,1}=tr(\boldsymbol{\beta}^{\mathrm{T}}\mathbf{U}\boldsymbol{\beta}) \tag{14}$$

where $\mathbf{U}$ is defined as Eq.(13).

Intuitively, $\mathbf{U}_{ii}$ becomes larger with the decreasing of $\|\boldsymbol{\beta}_i\|_2$, and the minimization of Eq.(11) tends to derive $\boldsymbol{\beta}_i$ with much smaller $\ell_2$-norm close to zero, *i.e.*, a sparse $\boldsymbol{\beta}$ is obtained. Note that if $\boldsymbol{\beta}_i=\mathbf{0}$, a very small value $\epsilon>0$ will be introduced, *i.e.*, $\|\boldsymbol{\beta}_i\|_2+\epsilon$, to update $\mathbf{U}$, then there is

$$\mathbf{U}_{ii}=\frac{1}{2(\|\boldsymbol{\beta}_i\|_2+\epsilon)},\epsilon>0 \tag{15}$$

The optimal $\boldsymbol{\beta}$ is solved by setting $\frac{\partial\mathcal{J}(\boldsymbol{\beta},\boldsymbol{\Theta})}{\partial\boldsymbol{\beta}}=0$, then we have

$$\boldsymbol{\beta}=\left(\mathbf{U}+C_{\mathcal{S}}\mathbf{H}_{\mathcal{S}}^{\mathrm{T}}\mathbf{H}_{\mathcal{S}}+C_{\mathcal{T}}\mathbf{H}_{\mathcal{T}}^{\mathrm{T}}\mathbf{H}_{\mathcal{T}}+\tau\mathbf{H}_{\mathcal{T}u}^{\mathrm{T}}\mathbf{H}_{\mathcal{T}u}+\lambda\mathbf{H}^{\mathrm{T}}\mathcal{L}\mathbf{H}\right)^{-1}(C_{\mathcal{S}}\mathbf{H}_{\mathcal{S}}^{\mathrm{T}}\mathbf{T}_{\mathcal{S}}+C_{\mathcal{T}}\mathbf{H}_{\mathcal{T}}^{\mathrm{T}}\mathbf{T}_{\mathcal{T}}\boldsymbol{\Theta}+\tau\mathbf{H}_{\mathcal{T}u}^{\mathrm{T}}\boldsymbol{\phi}_{p,\mathcal{T}u}) \tag{16}$$

Second, when $\boldsymbol{\beta}$ is fixed in one iteration, the optimization problem (11) becomes

---

**Algorithm 1.** Extreme Learning Machine based Domain Adaptation (EDA)

1: **Input:** $\mathbf{H}_{\mathcal{S}}$, $\mathbf{H}_{\mathcal{T}}$, $\mathbf{H}_{\mathcal{T}u}$, $\mathbf{H}$, $\mathbf{T}_{\mathcal{S}}$, $\mathbf{T}_{\mathcal{T}}$, $\boldsymbol{\phi}_p$, and $\mathcal{L}$;
2: **Initialization:** $\mathbf{U}^t\leftarrow\mathbf{I}_{L\times L}$, $\boldsymbol{\Theta}^t\leftarrow\mathbf{I}_{c\times c}$, $t\leftarrow1$;
3: **While** not converged $(t<T_{max})$ **do**
4:     Calculate the output weights $\boldsymbol{\beta}^t$ using (16);
5:     Update $\boldsymbol{\Theta}^{t+1}$ using (19);
6:     Update $\mathbf{U}^{t+1}$ using (15);
7:     $t\leftarrow t+1$;
8: **Until** convergence;
9: **Output:** $\boldsymbol{\beta}^t$ and $\boldsymbol{\Theta}^t$

---

$$\min_{\boldsymbol{\Theta}}\mathcal{J}(\boldsymbol{\Theta})=C_{\mathcal{T}}\|\mathbf{H}_{\mathcal{T}}\boldsymbol{\beta}-\mathbf{T}_{\mathcal{T}}\circ\boldsymbol{\Theta}\|_{\mathrm{F}}^2+\gamma\|\boldsymbol{\Theta}-\mathbf{I}\|_{\mathrm{F}}^2 \tag{17}$$

Then, one can update $\boldsymbol{\Theta}$ by setting the derivative of the objective function (17) *w.r.t.* $\boldsymbol{\Theta}$ to be zero. There is

$$\frac{\partial\mathcal{J}(\boldsymbol{\beta},\boldsymbol{\Theta})}{\partial\boldsymbol{\Theta}}=-2C_{\mathcal{T}}\mathbf{T}_{\mathcal{T}}^{\mathrm{T}}\mathbf{H}_{\mathcal{T}}\boldsymbol{\beta}+2C_{\mathcal{T}}\mathbf{T}_{\mathcal{T}}^{\mathrm{T}}\mathbf{T}_{\mathcal{T}}\boldsymbol{\Theta}+2\gamma\boldsymbol{\Theta}-2\gamma\mathbf{I}=\mathbf{0} \tag{18}$$

From (18), we can obtain the expression of $\boldsymbol{\Theta}$ as follows

$$\boldsymbol{\Theta}=\left(C_{\mathcal{T}}\mathbf{T}_{\mathcal{T}}^{\mathrm{T}}\mathbf{T}_{\mathcal{T}}+\gamma\mathbf{I}\right)^{-1}\left(C_{\mathcal{T}}\mathbf{T}_{\mathcal{T}}^{\mathrm{T}}\mathbf{H}_{\mathcal{T}}\boldsymbol{\beta}+\gamma\mathbf{I}\right) \tag{19}$$

The whole optimization procedure of EDA is summarized in Algorithm 1. The iterative update procedure is terminated once the number of iterations reaches $T_{max}$. In terms of experiments and convergence, we set $T_{max}$ to be 5 in this work. From Algorithm 1, we observe that the two variables are iteratively solved in closed-form with computational complexity of $\mathcal{O}(L^2)$ and $\mathcal{O}(c^2)$ using (20) and (19), respectively.

### D. Convergence Analysis of EDA

Since EDA is solved in an alternative way, its convergence behavior should be discussed. First, two lemmas are provided.

**Lemma 1 ([49]).** *For any non-zero vectors $\boldsymbol{a},\boldsymbol{b}\in\mathbb{R}^c$, there is*

$$\|\boldsymbol{a}\|_2-\frac{\|\boldsymbol{a}\|_2^2}{2\|\boldsymbol{b}\|_2}\leq\|\boldsymbol{b}\|_2-\frac{\|\boldsymbol{b}\|_2^2}{2\|\boldsymbol{b}\|_2} \tag{20}$$

Under the Lemma 1, we have the following Lemma 2.

**Lemma 2.** *For alternative optimization, after fixing $\mathbf{U}^t$ the two steps that when fix $\boldsymbol{\Theta}^t$, update $\boldsymbol{\beta}^t$, and when fix $\boldsymbol{\beta}^t$, update $\boldsymbol{\Theta}^{t+1}$ will not increase the complete joint objective function (11). Two claims with proofs are given as follows:*
*Claim 1.* $\mathcal{J}(\boldsymbol{\beta}^t,\boldsymbol{\Theta}^t,\mathbf{U}^t)\geq\mathcal{J}(\boldsymbol{\beta}^{t+1},\boldsymbol{\Theta}^t,\mathbf{U}^{t+1})$
*Claim 2.* $\mathcal{J}(\boldsymbol{\beta}^{t+1},\boldsymbol{\Theta}^t,\mathbf{U}^{t+1})\geq\mathcal{J}(\boldsymbol{\beta}^{t+1},\boldsymbol{\Theta}^{t+1},\mathbf{U}^{t+1})$
The proofs of *claim 1* and *2* are shown in **Supplementary Material**. The EDA convergence is summarized as theorem 1.

**Theorem 1.** *The joint objective function in (11) is monotonically non-increasing by employing the optimization procedure in Algorithm 1.*
*Proof.* As can be seen from *claim 1* and *claim 2* in Lemma 2, it is easy to obtain that the complete joint function will converge from one iteration to the next, and there is

$$\mathcal{J}(\boldsymbol{\beta}^t,\boldsymbol{\Theta}^t,\mathbf{U}^t)\geq\mathcal{J}(\boldsymbol{\beta}^{t+1},\boldsymbol{\Theta}^t,\mathbf{U}^{t+1})\geq\mathcal{J}(\boldsymbol{\beta}^{t+1},\boldsymbol{\Theta}^{t+1},\mathbf{U}^{t+1}) \tag{21}$$

Notably, due to that $\mathbf{U}$ is not dominant in the objective function (11) and $\mathbf{U}$ is just an intermediate variable which is completely determined when $\boldsymbol{\beta}$ is fixed as shown in (15), we have $\mathcal{J}(\boldsymbol{\beta}^t,\boldsymbol{\Theta}^t)\geq\mathcal{J}(\boldsymbol{\beta}^{t+1},\boldsymbol{\Theta}^{t+1})$. Then, **Theorem 1** is proven. One point should be denoted that the above theorem only indicates that the objective function is non-increasing. The convergence of $\boldsymbol{\beta}$ can be observed by calculating the difference $\|\boldsymbol{\beta}^t-\boldsymbol{\beta}^{t-1}\|_{\mathrm{F}}$. Additionally, the objective function based on $\ell_{2,1}$-norm is convex and the closed-form solution of each

variable is calculated. Therefore, the algorithm can converge to a global optimum after several iterations.

## IV. EDA OF MULTIPLE VIEWS (MVEDA)

In this section, motivated by the multi-feature learning as well as multi-view learning, we exploit the EDA in multiple *views* and induce a new method (*i.e.,* MvEDA), which is an extension of EDA. It is used to address the scenario where images/videos are represented with multiple features. MvEDA does not simply combine features (*i.e.,* feature concatenation), but fully exploits the complementary structural information and correlation among multiple features, by simultaneously learning an intrinsic data manifold for each feature.

### A. Multi-view EDA (MvEDA)

The proposed EDA of $V$ views based on (9) and (10) is formulated as follows

$$\min_{\boldsymbol{\beta}_v,\boldsymbol{\Theta}_v,\alpha_v,\boldsymbol{\xi}_{\mathcal{S},v}^i,\boldsymbol{\xi}_{\mathcal{T},v}^j,\boldsymbol{\xi}_{\mathcal{T}u,v}^k} \Sigma_{v=1}^V \|\boldsymbol{\beta}_v\|_{2,1} + C_{\mathcal{S}}\Sigma_{i=1}^{N_{\mathcal{S}\ell}}\Sigma_{v=1}^V \alpha_v\|\boldsymbol{\xi}_{\mathcal{S},v}^i\|^2 +$$
$$C_{\mathcal{T}}\Sigma_{j=1}^{N_{\mathcal{T}\ell}}\Sigma_{v=1}^V \alpha_v\|\boldsymbol{\xi}_{\mathcal{T},v}^j|_{\boldsymbol{\beta}_v,\boldsymbol{\Theta}_v}\|^2 + \gamma\Sigma_{v=1}^V \alpha_v\|\boldsymbol{\Theta}_v - \mathbf{I}\|_F^2 +$$
$$\tau\Sigma_{k=1}^{N_{\mathcal{T}u}}\Sigma_{v=1}^V \alpha_v \|\boldsymbol{\xi}_{\mathcal{T}u,v}^k|_{\boldsymbol{\beta}_v,\boldsymbol{\Phi}_p^v}\|^2 + \lambda \cdot tr(\Sigma_{v=1}^V \alpha_v^r \mathbf{F}_v^{\mathrm{T}}\mathcal{L}_v \mathbf{F}_v) \quad (22)$$

$$\text{s.t.} \begin{cases} \mathbf{H}_{\mathcal{S},v}^i\boldsymbol{\beta}_v = \mathbf{t}_{\mathcal{S}}^i - \boldsymbol{\xi}_{\mathcal{S},v}^i, i=1,\dots,N_{\mathcal{S}\ell} \\ \mathbf{H}_{\mathcal{T},v}^j\boldsymbol{\beta}_v = (\mathbf{t}_{\mathcal{T}}^j)^{\mathrm{T}}\circ\boldsymbol{\Theta}_v - \boldsymbol{\xi}_{\mathcal{T},v}^j|_{\boldsymbol{\beta}_v,\boldsymbol{\Theta}_v}, j=1,\dots,N_{\mathcal{T}\ell} \\ \mathbf{H}_{\mathcal{T}u,v}^k\boldsymbol{\beta}_v = \boldsymbol{\phi}_{p,\mathcal{T}u}^{k,v} - \boldsymbol{\xi}_{\mathcal{T}u,v}^k|_{\boldsymbol{\beta}_v,\boldsymbol{\Phi}_p^v}, k=1,\dots,N_{\mathcal{T}u} \\ \mathbf{F}_v = \mathbf{H}_v\boldsymbol{\beta}_v \\ \Sigma_{v=1}^V \alpha_v = 1, 0 < \alpha_v < 1 \\ r > 1, \gamma, \lambda, \tau, C_{\mathcal{S}}, C_{\mathcal{T}} > 0 \end{cases} \quad (23)$$

where $\alpha_v$ represents the weighted coefficient of the $v$-th feature. From the expression (22), MvEDA belongs to a multi-view learning framework, which fully exploits the manifold structure of the intrinsic data geometry implied in multiple features, by learning the coefficient $\alpha_v$ and the feature specific classifier $\boldsymbol{\beta}_v$ *w.r.t.* the $v$-th feature. Therefore, the underlying feature correlation and complementary structural information among multiple features can be exploited during domain adaptation, which results in a more robust classification in complex data.

Similar to formulation of EDA, by substituting the constraints (23) into the objective function (22), the MvEDA can be reformulated as

$$\mathcal{J}(\boldsymbol{\beta}_v^g,\boldsymbol{\Theta}_v,\alpha_v) = \Sigma_{v=1}^V \|\boldsymbol{\beta}_v\|_{2,1} + C_{\mathcal{S}}\Sigma_{i=1}^{N_{\mathcal{S}}}\Sigma_{v=1}^V \alpha_v\|\mathbf{H}_{\mathcal{S},v}^i\boldsymbol{\beta}_v - \mathbf{t}_{\mathcal{S}}^i\|_2^2 +$$
$$C_{\mathcal{T}}\Sigma_{j=1}^{N_{\mathcal{T}}}\Sigma_{v=1}^V \alpha_v\|\mathbf{H}_{\mathcal{T},v}^j\boldsymbol{\beta}_v - \mathbf{t}_{\mathcal{T}}^j\circ\boldsymbol{\Theta}_v\|_2^2 + \gamma\Sigma_{v=1}^V \alpha_v\|\boldsymbol{\Theta}_v - \mathbf{I}\|_F^2 +$$
$$\tau\Sigma_{k=1}^{N_{\mathcal{T}u}}\Sigma_{v=1}^V \alpha_v\|\mathbf{H}_{\mathcal{T}u,v}^k\boldsymbol{\beta}_v - \boldsymbol{\phi}_{p,\mathcal{T}u}^{k,v}\|^2 + \lambda tr(\Sigma_{v=1}^V \alpha_v^r \mathbf{F}_v^{\mathrm{T}}\mathcal{L}_v \mathbf{F}_v) \quad (24)$$

which can be compactly rewritten as

$$\mathcal{J}(\boldsymbol{\beta}_v^g,\boldsymbol{\Theta}_v,\alpha_v) = \Sigma_{v=1}^V \|\boldsymbol{\beta}_v\|_{2,1} + C_{\mathcal{S}}\Sigma_{v=1}^V \alpha_v\|\mathbf{H}_{\mathcal{S},v}\boldsymbol{\beta}_v - \mathbf{T}_{\mathcal{S}}\|_F^2 +$$
$$C_{\mathcal{T}}\Sigma_{v=1}^V \alpha_v\|\mathbf{H}_{\mathcal{T},v}\boldsymbol{\beta}_v - \mathbf{T}_{\mathcal{T}}\circ\boldsymbol{\Theta}_v\|_F^2 + \gamma\Sigma_{v=1}^V \alpha_v\|\boldsymbol{\Theta}_v - \mathbf{I}\|_F^2 +$$
$$\tau\Sigma_{v=1}^V \alpha_v\|\mathbf{H}_{\mathcal{T}u,v}\boldsymbol{\beta}_v - \boldsymbol{\Phi}_{p,\mathcal{T}u}^v\|_F^2 + \lambda tr(\Sigma_{v=1}^V \boldsymbol{\beta}_v^{\mathrm{T}}\mathbf{H}_v^{\mathrm{T}}\alpha_v^r\mathcal{L}_v\mathbf{H}_v\boldsymbol{\beta}_v) \quad (25)$$

Note that the setting of $r>1$ of $\alpha_v^r$ is to better exploit the complementary structure information of multiple features, and avoid that case where only the best modality is considered (*e.g.,* $\alpha_v = 1$). In this paper, we set $r$=2 in experiment.

### B. Optimization Algorithm

As can be seen from the objective function (25), it is convex *w.r.t.* $\boldsymbol{\beta}_v$ when $\boldsymbol{\Theta}_v$ and $\alpha_v$ are fixed. Note that $\|\boldsymbol{\beta}_v\|_{2,1}$ is convex, however, its derivative does not exist when $\boldsymbol{\beta}_v^i = \mathbf{0}$ for

$i=1,\dots,L$. Thus, when $\boldsymbol{\beta}_v^i \neq \mathbf{0}$ for $i=1,\dots,L$, by calculating the derivate of the objective function (25) *w.r.t.* $\boldsymbol{\beta}_v$, one can obtain

$$\frac{\partial\mathcal{J}(\boldsymbol{\beta}_v,\boldsymbol{\Theta}_v,\alpha_v)}{\partial\boldsymbol{\beta}_v} =$$
$$2(\mathbf{U}_v + C_{\mathcal{S}}\alpha_v\mathbf{H}_{\mathcal{S},v}^{\mathrm{T}}\mathbf{H}_{\mathcal{S},v} + C_{\mathcal{T}}\alpha_v\mathbf{H}_{\mathcal{T},v}^{\mathrm{T}}\mathbf{H}_{\mathcal{T},v} + \tau\alpha_v\mathbf{H}_{\mathcal{T}u,v}^{\mathrm{T}}\mathbf{H}_{\mathcal{T}u,v} +$$
$$\lambda\alpha_v^r\mathbf{H}_v^{\mathrm{T}}\mathcal{L}_v\mathbf{H}_v)\boldsymbol{\beta}_v - 2(C_{\mathcal{S}}\alpha_v\mathbf{H}_{\mathcal{S},v}^{\mathrm{T}}\mathbf{T}_{\mathcal{S}} + C_{\mathcal{T}}\alpha_v\mathbf{H}_{\mathcal{T},v}^{\mathrm{T}}\mathbf{T}_{\mathcal{T}}\boldsymbol{\Theta}_v +$$
$$\tau\alpha_v\mathbf{H}_{\mathcal{T}u,v}^{\mathrm{T}}\boldsymbol{\Phi}_{p,\mathcal{T}u}^v) \quad (26)$$

where $\mathbf{U}_v \in \mathbb{R}^{L\times L}$ is formulated similarly with (13).

*First*, fix $\alpha_v = 1/V$ and $\boldsymbol{\Theta}_v = \mathbf{I}_{c\times c}$, let $\frac{\partial\mathcal{J}(\boldsymbol{\beta}_v,\boldsymbol{\alpha},\boldsymbol{\Theta})}{\partial\boldsymbol{\beta}_v} = 0$, then $\boldsymbol{\beta}_v$ can be solved as

$$\boldsymbol{\beta}_v = (\mathbf{U}_v + C_{\mathcal{S}}\alpha_v\mathbf{H}_{\mathcal{S},v}^{\mathrm{T}}\mathbf{H}_{\mathcal{S},v} + C_{\mathcal{T}}\alpha_v\mathbf{H}_{\mathcal{T},v}^{\mathrm{T}}\mathbf{H}_{\mathcal{T},v} + \tau\alpha_v\mathbf{H}_{\mathcal{T}u,v}^{\mathrm{T}}\mathbf{H}_{\mathcal{T}u,v} +$$
$$\lambda\alpha_v^r\mathbf{H}_v^{\mathrm{T}}\mathcal{L}_v\mathbf{H}_v)^{-1}(C_{\mathcal{S}}\alpha_v\mathbf{H}_{\mathcal{S},v}^{\mathrm{T}}\mathbf{T}_{\mathcal{S}} + C_{\mathcal{T}}\alpha_v\mathbf{H}_{\mathcal{T},v}^{\mathrm{T}}\mathbf{T}_{\mathcal{T}}\boldsymbol{\Theta}_v + \tau\alpha_v\mathbf{H}_{\mathcal{T}u,v}^{\mathrm{T}}\boldsymbol{\Phi}_{p,\mathcal{T}u}^v) \quad (27)$$

Once $\boldsymbol{\beta}_v$ is solved, $\mathbf{U}_v$ can be intuitively calculated by adding a small perturbation $\epsilon$ as shown in (15).

*Second*, when $\boldsymbol{\beta}_v$ is fixed, the optimization becomes

$$\min_{\boldsymbol{\Theta}_v} C_{\mathcal{T}}\Sigma_{v=1}^V \alpha_v\|\mathbf{H}_{\mathcal{T},v}\boldsymbol{\beta}_v - \mathbf{T}_{\mathcal{T}}\circ\boldsymbol{\Theta}_v\|_F^2 + \gamma\Sigma_{v=1}^V \alpha_v\|\boldsymbol{\Theta}_v - \mathbf{I}\|_F^2 \quad (28)$$

The objective function (28) is convex *w.r.t.* $\boldsymbol{\Theta}_v$. By setting $\frac{\partial\mathcal{J}(\boldsymbol{\beta}_v,\boldsymbol{\Theta}_v,\alpha_v)}{\partial\boldsymbol{\Theta}_v} = 0$, the update rule of $\boldsymbol{\Theta}_v$ is obtained as follows

$$\boldsymbol{\Theta}_v = (C_{\mathcal{T}}\alpha_v\mathbf{T}_{\mathcal{T}}^{\mathrm{T}}\mathbf{T}_{\mathcal{T}} + \gamma\alpha_v\mathbf{I})^{-1}(C_{\mathcal{T}}\alpha_v\mathbf{T}_{\mathcal{T}}^{\mathrm{T}}\mathbf{H}_{\mathcal{T},v}\boldsymbol{\beta}_v + \gamma\alpha_v\mathbf{I}) \quad (29)$$

*Third*, after fixing $\boldsymbol{\beta}_v$ and $\boldsymbol{\Theta}_v$, we update $\alpha_v$ using Lagrange multiplier method with constraints, which is formulated as

$$\mathcal{J}(\boldsymbol{\beta}_v,\boldsymbol{\Theta}_v,\alpha_v,\eta) = \Sigma_{v=1}^V \|\boldsymbol{\beta}_v\|_{2,1} + C_{\mathcal{S}}\Sigma_{v=1}^V \alpha_v\|\mathbf{H}_{\mathcal{S},v}\boldsymbol{\beta}_v - \mathbf{T}_{\mathcal{S}}\|_F^2 +$$
$$C_{\mathcal{T}}\Sigma_{v=1}^V \alpha_v\|\mathbf{H}_{\mathcal{T},v}\boldsymbol{\beta}_v - \mathbf{T}_{\mathcal{T}}\circ\boldsymbol{\Theta}_v\|_F^2 + \gamma\Sigma_{v=1}^V \alpha_v\|\boldsymbol{\Theta}_v - \mathbf{I}\|_F^2 +$$
$$\tau\Sigma_{v=1}^V \alpha_v\|\mathbf{H}_{\mathcal{T}u,v}\boldsymbol{\beta}_v - \boldsymbol{\Phi}_{p,\mathcal{T}u}^v\|_F^2 + \lambda\cdot tr(\Sigma_{v=1}^V \boldsymbol{\beta}_v^{\mathrm{T}}\mathbf{H}_v^{\mathrm{T}}\alpha_v^r\mathcal{L}_v\mathbf{H}_v\boldsymbol{\beta}_v) -$$
$$\eta\cdot(\Sigma_{v=1}^V \alpha_v - 1) \quad (30)$$

where $\eta$ denotes the Lagrange multiplier.

By calculating the derivatives of (30) *w.r.t.* $\alpha_v$ and $\eta$, one can obtain the following equations

$$\begin{cases} \frac{\partial\mathcal{J}(\boldsymbol{\beta}_v,\alpha_v,\eta)}{\partial\alpha_v} = 0 \rightarrow r\alpha_v^{r-1}tr(\boldsymbol{\beta}_v^{\mathrm{T}}\mathbf{H}_v^{\mathrm{T}}\mathcal{L}_v\mathbf{H}_v\boldsymbol{\beta}_v) + \mathcal{G} = 0 \\ \frac{\partial\mathcal{J}(\boldsymbol{\beta}_v,\alpha_v,\eta)}{\partial\eta} = 0 \rightarrow \Sigma_{v=1}^V \alpha_v - 1 = 0 \end{cases} \quad (31)$$

where the variable $\mathcal{G}$ is calculated as follow

$$\mathcal{G} = 1/\lambda\Big(C_{\mathcal{S}}\|\mathbf{H}_{\mathcal{S},v}\boldsymbol{\beta}_v - \mathbf{T}_{\mathcal{S}}\|_F^2 + C_{\mathcal{T}}\|\mathbf{H}_{\mathcal{T},v}\boldsymbol{\beta}_v - \mathbf{T}_{\mathcal{T}}\circ\boldsymbol{\Theta}_v\|_F^2 +$$
$$\gamma\|\boldsymbol{\Theta}_v - \mathbf{I}\|_F^2 + \tau\|\mathbf{H}_{\mathcal{T}u,v}\boldsymbol{\beta}_v - \boldsymbol{\Phi}_{p,\mathcal{T}u}^v\|_F^2 - \eta\Big) \quad (32)$$

By solving the Eq.(31), it is easy to get the expression of $\alpha_v$ as

$$\alpha_v = \Big(\frac{1}{tr(\boldsymbol{\beta}_v^{\mathrm{T}}\mathbf{H}_v^{\mathrm{T}}\mathcal{L}_v\mathbf{H}_v\boldsymbol{\beta}_v)}\Big)^{\frac{1}{r-1}}\Big/\Sigma_{v=1}^V \Big(\frac{1}{tr(\boldsymbol{\beta}_v^{\mathrm{T}}\mathbf{H}_v^{\mathrm{T}}\mathcal{L}_v\mathbf{H}_v\boldsymbol{\beta}_v)}\Big)^{\frac{1}{r-1}} \quad (33)$$

where $r$ ($r>1$) is set as 2 in this paper. In summary, an efficient alternative optimization is presented in Algorithm 2 to solve the proposed MvEDA model.

From the viewpoint of Algorithm 1 in single view, we can see that it is a special case of Algorithm 2 when $V$=1. In this paper, we call the proposed unified framework as EDA. It is worth noting that although an alternative optimization is used in EDA, the proposed EDA is still accord with the conventional ELM framework in computing the output weights $\boldsymbol{\beta}$, into which three steps are incorporated: 1) ELM network (SLFN) initialization; 2) feature mapping in hidden layer with randomly generated input weights and bias; 3) analytically determine the output weights $\beta$ (*i.e.*, closed-form solution). The complete training algorithm of the unified EDA framework for implementation is summarized as Algorithm 3.

**Algorithm 2.** Multi-view EDA
1: **Input:** $\mathbf{H}_{\mathcal{S}}^v, \mathbf{H}_{\mathcal{T}}^v, \mathbf{H}_{\mathcal{T}u}^v, \mathbf{H}^v, \mathbf{T}_{\mathcal{S}}, \mathbf{T}_{\mathcal{T}}, \Phi_{\mathcal{P}}^v, \mathcal{L}_v;$
2: **Initialization:** $\mathbf{U}_v^t \leftarrow \mathbf{I}_{L\times L}, \Theta_v^t \leftarrow \mathbf{I}_{c\times c}, \alpha_v^t \leftarrow 1/V, t \leftarrow 1;$
3: **While** not converged $(t < T_{max})$ **do**
4:    Calculate the output weights $\beta_v^t$ using (27);
5:    update $\Theta_v^{t+1}$ using (29);
6:    update $\alpha_v^{t+1}$ using (33);
7:    update $\mathbf{U}_v^{t+1}$ using (15);
8:    $t \leftarrow t + 1;$
9: **Until** Convergence;
10: **Output:** $\beta_v, \Theta_v$ and $\alpha_v$ $(v=1,\dots,V)$

### C. Convergence analysis of MvEDA

To explore the convergence analysis of the proposed MvEDA framework shown in Algorithm 2, we first provide a lemma 3 as follows

**Lemma 3.** *For alternative optimization, after fixing $\mathbf{U}_v^t$ the three update steps that update $\beta_v^t$, update $\Theta_v^{t+1}$ and update $\alpha_v^{t+1}$ will not increase the complete joint objective function. Three claims are given:*

*Claim 3.* $\mathcal{J}(\beta_v^t, \Theta_v^t, \alpha_v^t, \mathbf{U}_v^t) \geq \mathcal{J}(\beta_v^{t+1}, \Theta_v^t, \alpha_v^t, \mathbf{U}_v^{t+1})$

*Claim 4.* $\mathcal{J}(\beta_v^{t+1}, \Theta_v^t, \alpha_v^t, \mathbf{U}_v^{t+1}) \geq \mathcal{J}(\beta_v^{t+1}, \Theta_v^{t+1}, \alpha_v^t, \mathbf{U}_v^{t+1})$

*Claim 5.* $\mathcal{J}(\beta_v^{t+1}, \Theta_v^{t+1}, \alpha_v^t, \mathbf{U}_v^{t+1}) \geq \mathcal{J}(\beta_v^{t+1}, \Theta_v^{t+1}, \alpha_v^{t+1}, \mathbf{U}_v^{t+1})$

The proofs of *claim 3*, *claim 4*, and *claim 5* are provided in **Supplementary Material**. Further, the convergence of MvEDA is summarized as the following theorem 2.

**Theorem 2:** *The joint objective function in (25) is monotonically non-increasing by employing the optimization procedure in Algorithm 2.*
*Proof.* As can be seen from *claim 3*, *claim 4*, and *claim 5* in Lemma 3, it is easy to obtain that the complete joint function will converge from one iteration to the next.

$\mathcal{J}(\beta_v^t, \Theta_v^t, \alpha_v^t) = \mathcal{J}(\beta_v^t, \Theta_v^t, \alpha_v^t, \mathbf{U}_v^t) \geq \mathcal{J}(\beta_v^{t+1}, \Theta_v^t, \alpha_v^t, \mathbf{U}_v^{t+1})$
$\geq \mathcal{J}(\beta_v^{t+1}, \Theta_v^{t+1}, \alpha_v^t, \mathbf{U}_v^{t+1}) \geq \mathcal{J}(\beta_v^{t+1}, \Theta_v^{t+1}, \alpha_v^{t+1}, \mathbf{U}_v^{t+1})$

Then **Theorem 2** is proven. Note that $\mathbf{U}_v^t$ is simultaneously determined when $\beta_v^t$ is fixed according to (15), thus $\mathbf{U}_v^{t+1}$ is also determined when $\beta_v^{t+1}$ is fixed.

## V. EXPERIMENTS

In this section, we evaluate our proposed methods EDA and MvEDA on four datasets: 1) the challenging YouTube & Consumer videos (SIFT and ST features), 2) the *3DA Office* dataset (SURF feature), 3) the *4DA Extended office* dataset (SURF vs. CNN features), 4) the Bing-Caltech dataset (*Classeme* feature). Notably, EDA is termed for single feature scenarios and MvEDA is termed for multiple features based application scenarios (*e.g.*, YouTube videos).

### A. Brief Descriptions of Experimental Datasets

■ **YouTube & Consumer Videos Dataset**

This dataset was developed for visual event recognition and evaluating semi-supervised domain adaption approaches, in which part of the consumer videos were derived from Kodak

**Algorithm 3.** Complete EDA
**Input:**
1: Training samples $\{\mathbf{X}_{\mathcal{S},v}, \mathbf{T}_{\mathcal{S},v}\} = \{\mathbf{x}_{\mathcal{S},v}^i, t_{\mathcal{S},v}^i\}_{i=1}^{N_S}$ of the source domain $\mathcal{S}$ *w.r.t.* the *v*-th modality, $v = 1, \cdots, V;$
2: Labeled guide samples $\{\mathbf{X}_{\mathcal{T},v}, \mathbf{T}_{\mathcal{T},v}\} = \{\mathbf{x}_{\mathcal{T},v}^j, t_{\mathcal{T},v}^j\}_{j=1}^{N_{\mathcal{T}l}}$ of the target domain $\mathcal{T}$ *w.r.t.* the *v*-th modality, $v = 1, \cdots, V;$
3: Unlabeled samples $\{\mathbf{X}_{\mathcal{T}u,v}, \mathbf{T}_{\mathcal{T}u,v}\} = \{\mathbf{x}_{\mathcal{T}u,v}^j, t_{\mathcal{T}u,v}^j\}_{j=1}^{N_{\mathcal{T}u}}$ of the target domain $\mathcal{T}$ *w.r.t.* the modality, $v = 1, \cdots, V;$
4: The trade-off parameters;
**Output:** $\beta_v$ and $\alpha_v$ $(v = 1, \cdots, V)$
**Procedure:**
*Stage 1. EDA Network Initialization.*
5: Initialize the EDA network of $L$ hidden neurons with randomly selected input weights $\mathbf{W}$ and hidden bias $\mathbf{B}$ with 0-1 uniform distribution.
*Stage 2. EDA Feature Mapping and Graph Construction.*
6: Calculate the hidden layer output matrix $\mathbf{H}_{\mathcal{S},v}, \mathbf{H}_{\mathcal{T},v}, \mathbf{H}_{\mathcal{T}u,v}$ as $\mathbf{H}_{\mathcal{S},v} = \mathcal{H}(\mathbf{W} \cdot \mathbf{X}_{\mathcal{S},v} + \mathbf{B})$, $\mathbf{H}_{\mathcal{T},v} = \mathcal{H}(\mathbf{W} \cdot \mathbf{X}_{\mathcal{T},v} + \mathbf{B})$ and $\mathbf{H}_{\mathcal{T}u,v} = \mathcal{H}(\mathbf{W} \cdot \mathbf{X}_{\mathcal{T}u,v} + \mathbf{B})$, respectively;
7: Compute $\mathbf{H}_v$ *w.r.t.* all the instances in target domain;
8: Compute the graph Laplacian matrix $\mathcal{L}_v;$
*Stage 3. Learning algorithm.*
9: **if** $V<2$ **then**
10:    Call Algorithm 1.
11: **else** Call Algorithm 2.
12: **end if**

Consumer Video Benchmark Data Set [5], and part of new consumer video clips from real users were collected by Duan *et al.* [19]. The YouTube videos were from the website[1]. Totally, six events such as "birthday", "picnic", "parade", "show", "sports" and "wedding" of 195 consumer videos (target domain) and 906 YouTube videos (source domain, *i.e.*, web videos) are included in the dataset. Notably the domain bias/shift results from camera viewpoint, resolution of imaging sensor, background, etc. as shown in Fig.1.

■ **3DA Dataset (office data)**

We employ the benchmark 3DA dataset in [37] for object recognition, which was introduced for evaluating visual domain adaptation approaches. This database shows a challenging office environment and reflects the difficult task of real-world object recognition. This dataset contains totally 4106 images of 31 categories from three domains:

*Amazon (object images from the web)*: The *amazon* domain contains 2813 images, which capture a large intra-class variation across categories and shows typical appearances.

*Webcam (low-resolution images by a webcam)*: The *webcam* domain contains 795 images, with five objects per class captured on five different viewpoints per object on average. The images are of low resolution, showing significant noise and visual domain shifts in a realistic office environment.

*Dslr (high-resolution images by a digital SLR camera)*: The *dslr* domain consists of 498 images from 31 classes, which are captured in realistic office environment under natural lighting condition by a digital SLR camera. The same five objects per category as the *webcam* were used, and three images per object taken from different camera viewpoints were captured on average. These images are of higher resolution and lower noise.

Note that the dimensionality of source data (*i.e.*, *webcam* and

---

1.http://vc.sce.ntu.edu.sg/transfer_learning_domain_adaptation/domain_adaptation_home.html

*amazon*) is 800 (*i.e.*, the length of each sample vector), while that of target data (*i.e.*, *dslr*) is 600. The SURF feature [37] was extracted for each domain.

■ **4DA Dataset (extended office data)**

This 4DA dataset [43] is an extended version of 3DA *office* dataset with an extra *caltech* domain, resulting in four domains: *amazon*, *webcam*, *dslr* and *caltech*. 10 categories sampled from the *3DA* data set consist of the first 3 domains, while the *caltech* domain is sampled from the well-known *Caltech256* data. For 4DA dataset, two kinds of features are exploited: 800bin-SURF feature [37], [43] and 4096-dims convolutional neural net (CNN) feature [52]. Specifically, the total number of samples is 958, 295, 157, and 1123 for *amazon*, *webcam*, *dslr*, and *caltech* domains, respectively. Notably, the structures of CNN trained on the ImageNet with 1000 categories are the same as the proposed CNN in [58], which includes 5 convolutional layers and 3 fully-connected layers. The well-trained network parameters are used for deep representation of the 4DA dataset. The CNN outputs of the $6^{th}$ ($f_6$) and $7^{th}$ ($f_7$) fully-connected layers would be used as the input features of the EDA model.

■ **Bing-Caltech Dataset**

To demonstrate the effect of our method on large-scale dataset, the *Bing* dataset from [38], which has a larger number of images per class than the *office* data, is used in this paper. The *Bing* data sampled by Bing search engine that contains 300 images per class (256 classes) is used as *source* domain. The *Caltech 256* data that contains 80 images per class is used as *target* domain. The *classeme* features with 2625 keywords computed by the developers of the dataset in [38] are used in this paper. The number of classes for both domains is 256 and the dimensionality of each feature vector is 2625.

### B. Experimental Setup

■ **Details for YouTube & Consumer Videos Dataset**

We follow the same experimental settings as [20], where 906 loosely labeled YouTube videos are used as labeled training data in source domain. Additionally, $m$ ($m$=1, 3, 5, 7, 10) consumer videos per event are selected as the labeled training videos in target domain, respectively. The remaining videos of consumer videos data are then viewed as the unlabeled test data in target domain (note that they are also used as unlabeled training data in a semi-supervised setting). We adopt the given 5 random train/test splits of the labeled training videos from target domain in [20], and the MAPs (mean average precision) are reported. The features provided in [20] include scale invariant feature transform (SIFT with level $L$=0 and $L$=1) and space-time (ST with $L$=0 and $L$=1) features [49]. Clearly, the training data consists of source data and a limited number of labeled target data. Specifically, three feature-specific cases are studied: (a) SIFT with $L$=0 and $L$=1; (b) ST with $L$=0 and $L$=1; (c) SIFT+ST with $L$=0 and $L$=1.

The comparisons of algorithms for video event recognition are briefly described in the following two parts:

• We first compare with the conventional ELM, *i.e.*, ELM_S (ELM trained on source domain), ELM_T (ELM trained on target domain), ELM_ST (ELM trained on both source and target domains) and SS-ELM (semi-supervised ELM) [31]. For the proposed EDA, $EDA_{LapSVM}$ (SVM with Laplacian kernel $K(i,j) = \exp\left(-\sqrt{\sigma}D(v_i, v_j)\right)$), $EDA_{ldSVM}$ (SVM with inverse distance kernel $K(i,j) = \frac{1}{\sqrt{\sigma}D(v_i, v_j)+1}$) and $EDA_{avg}$ (i.e. $0.5(f_{LapSVM} + f_{ldSVM})$) are incorporated as pre-learned base classifiers, where $D(v_i, v_j)$ denotes the distance between the $i$-th and the $j$-th video, and $\sigma$ denotes the default kernel parameter $\frac{1}{A}$ ($A$ is the average of the square distance) [20].

• We then compare with the baseline methods including SVM_T (*i.e.*, SVM trained on target domain), SVM_ST (*i.e.*, SVM trained on both source and target domains), and MKL [30]. Additionally, the state-of-the-art domain adaptation approaches including FR [35], DTMKL [40] and AMKL [20] are also explored. For the proposed EDA, AMKL is shown as a pre-classifier in learning, *i.e.*, $EDA_{AMKL}$.

■ **Details for Office Dataset (3DA)**

For this dataset, the *amazon* and *webcam* domains are used as source domain, respectively, and the *dslr* domain is considered as target domain. We strictly follow the experimental setting in [21], [37]. To this end, 800-dimensional histogram features were obtained for *amazon* and *webcam* domains, and 600-dimensional histogram features were obtained for *dslr* domain through $k$-means cluster and vector quantization. The number of labeled data per class used is 20, 8, and 3, respectively for *amazon* (source), *webcam* (source) and *dslr* (target) domains.

The comparison algorithms are SVM_T, SGF [57], GFK [43], HeMap [41], DAMA [42], ARC-t [36], T-SVM [38], HFA [39] and SHFA [21]. The brief introduction of these methods can be referred in Section 2. The SVM with RBF kernel is used as pre-classifier in EDA, *i.e.*, $EDA_{SVM}$.

■ **Details for Extended Office Dataset (4DA)**

The 4DA dataset is from [43]. The same *SURF* features[2] that are vector quantized to 800 dimensions for *amazon*, *webcam*, *dslr*, and *caltech* domains are used. From the website[3], 20 kinds of train/test splits of the labeled samples from source and target domains can be obtained, and the average results across these splits are reported. We strictly follow the experimental settings in [23], [43]. The number of labeled source samples per class for *amazon*, *webcam*, *dslr* and *caltech* is 20, 8, 8, and 8, respectively, when they are used as source domain. Instead, when they are used as target domain, the number of labeled target samples per class is 3 for each domain. Additionally, the 4096-dimensional deep *CNN features* for the same dataset with completely the same experimental configuration have also been explored in experiments. Specifically, the images of the 4DA data are feed into the well-trained CNN, and the outputs of the $6^{th}$ ($f_6$) and $7^{th}$ ($f_7$) fully-connected layers are recognized to be the deep feature representations of the 4DA dataset, separately.

The compared algorithms for this dataset are SVM_S, SVM_T, SGF [57], GFK [43], HFA [39], ARC-t [36], Symm

---

[22], MMDT [22] and LandMark [55]. Their brief introductions can be captured in Section 2. For our EDA, SVM is used as a pre-classifier, *i.e.*, EDA$_{SVM}$.

■ **Details for Bing-Caltech Dataset**

For this dataset, with the *Bing* as source domain and *Caltech 256* as target domain, we follow the two settings as [38], [22]:

• Setting 1 [38]: fix $N_t$=10 target training samples per class, and vary the number $N_s$=5, 10, 50, 100, and 150 of source training samples per class, respectively.

• Setting 2 [22]: fix $N_s$=50 source training samples per class, and vary the number $N_t$=5, 10, 15, and 20 of target training samples per class, respectively. Note that only the data of the first 20 categories are considered in this setting.

*C. Parameter Tuning*

As can be seen from the proposed EDA model (33), five dominant model parameters such as $C_S$, $C_T$, $\gamma$, $\tau$, and $\lambda$ are referred. It's known to us that tuning five parameters are generally not robust for a machine learning algorithm. Therefore, it is unnecessary to tune all of them in real applications. In this work, we only free two important parameters with other parameters frozen. Intuitively, we set $\gamma = \lambda = 1$. For better exploiting the unlabeled data of target domain, a slightly larger coefficient $\tau$ can be used, *e.g.*, $\tau = 1000$. In experiments, cross-validation may not be suitable to tune the two free parameters $C_S$ and $C_T$ due to the limited number of labeled data in target domain. Therefore, following the same strategy [40] published in T-PAMI, the best result after tuning the two parameters from $\{1, 10, 10^2, 10^3, 10^4\}$ is reported. The parameter sensitivity has also been discussed in Section V, which indicates that the parameters can be empirically determined for obtaining an acceptable result. Additionally, in terms of ELM theory, the number of hidden nodes is not necessarily related with the performance. Empirically, we fix the number $L$ of hidden nodes in EDA approximately as $D\log N$. The hidden layer function is *radbas*.

*D. Experimental Results and Comparisons*

■ **Results on YouTube & Consumer Videos Dataset**

We first compare our EDA framework with ELM based methods for video event recognition. The comparison results with ELM based methods by using 3 labeled training samples per category from target domain are shown in Table I, from which we have the following observations:

• The ELM_ST trained on both domains achieves better results than ELM_S and ELM_T, which demonstrates that ELM can be explored for domain adaptation. SS-ELM [31] incorporates the manifold regularization to leverage unlabeled data, which achieves similar results with ELM_ST. The potential of ELM for cross-domain learning is shown.

• The proposed EDA is superior to the ELM based methods (10%, 3.7% and 12.7% improvements for three cases). This strongly proves that the proposed EDA has made an excellent contribution to ELM theory in domain adaptation.

• EDA with Laplacian kernel SVM, inverse distance kernel SVM and their average decision $\frac{1}{2}(f_{LapSVM} + f_{IdSVM})$ achieve

slightly different results, which also demonstrate that the pre-classifier for the unlabeled data in EDA is user and task-specific, and can be selected appropriately. One merit of EDA is that it can incorporate many different classifiers and becomes controllable as well as operable.

Furthermore, we compare our proposed method with baselines and state-of-the-art approaches, with $m$ ($m$=1, 3, 5, 7, 10) labeled training samples per class from the target domain. The results are reported in Table II. We can observe that,

• SVM_ST is better than SVM_T for case (a). It demonstrates that the classifier trained on a limited number of labeled training samples from target domain is not effective, by comparing to that classifier trained on both source and target data. However, we can also observe that SVM_ST is always worse than SVM_T for case (b) and (c). A potential and possible explanation is that the distribution of ST features between source domain and target domain is not dense, such that the source data has negative effect on the performance when ST features are used in case (b) and (c).

• For all methods, the recognition performance learned on SIFT features (case a) is better than that based on ST features (case b). This demonstrates that SIFT features are more robust for classifier learning. For DTMKL, A-MKL and the proposed EDA, the combined case c (SIFT+ST) is always better than that of case a. This demonstrates that information sharing and fusion among multi-features can promote the classification. However, for other methods under case c, the performance is not obvious. The reason is that the correspondence between two kinds of features is not bridged, that is, simple concatenation does not well analyze features. Comparatively, the A-MKL and our method can learn optimal weights to realize the fusion of SIFT and ST features. Additionally, the proposed MvEDA can deeply insight the features better with a joint multi-view learning framework.

• AMKL is better than DTMKL for all cases, which is accord with [20]. DTMKL is always worse than MKL for case (b). However, A-MKL is worse than MKL for case (b) when $m$=7 and 10. One possible explanation is that ST features distribute sparsely in feature space. Learning only on very few number of target data has weak domain adaptability.

• Both SVM_ST and FR are always better than MKL for case (a). This proves that cross-domain learning on dense SIFT features of both domains can achieve better performance than those without considering domain adaptation.

• Our proposed EDA achieves the best performance by constructing a single layer feed-forward network, trained with semi-supervised cross-domain learning algorithms as well as multiple views. We also prove that the pre-classifiers for exploiting the unlabeled data in target domain can make the learned EDA classifier more stable. The manifold regularization can also improve the recognition ability of the unlabeled data. We have highlighted the best, second best results and the increments in Table II, from which, we can observe that the proposed EDA has a significant improvement than A-MKL (4.7% improvement at the most). However, for case (b) with $m$=7 and 10, the A-MKL is much worse than EDA (8.8% and 6.0% improvement).

TABLE I
MEANS AND STANDARD DEVIATIONS OF MAPs (%) IN THREE CASES

| Methods | ELM_S | ELM_T | ELM_ST | SS-ELM | EDA$_{LapSVM}$ | EDA$_{tsSVM}$ | EDA$_{avg}$ | Improvement |
|---|---|---|---|---|---|---|---|---|
| SIFT | 38.9±2.5 | 36.0±3.6 | 40.3±2.2 | **40.3±2.2** | 49.0±1.9 | 50.1±1.4 | **50.3±2.4** | 10.0% |
| ST | 32.8±2.4 | 26.0±1.9 | 33.5±1.2 | **33.5±1.2** | 36.2±1.7 | 36.6±1.6 | **37.2±2.1** | 3.7% |
| Methods | ELM_S | ELM_T | ELM_ST | SS-ELM | MvEDA$_{LapSVM}$ | MvEDA$_{tsSVM}$ | MvEDA$_{avg}$ | Improvement |
| SIFT+ST | 43.3±2.0 | 36.4±5.5 | 44.4±1.7 | **44.4±1.6** | 55.6±1.8 | **57.1±2.2** | 56.3±1.9 | 12.7% |

TABLE II
MEAN AVERAGE PRECISION (%) OF THREE CASES WITH DIFFERENT NUMBER OF LABELED TARGET TRAINING DATA ($M$=1, 3, 5, 7, 10)

| # $m$ | Methods | SVM_T | SVM_ST | FR | MKL | DTMKL | A-MKL | EDA$_{AMKL}$ | MvEDA | Improvement |
|---|---|---|---|---|---|---|---|---|---|---|
| 1 | SIFT | 38.8±4.8 | 49.4±3.2 | 47.3±0.4 | 43.9±2.4 | 48.7±1.4 | 51.6±1.4 | **51.9±1.0** | - | 0.3% |
|   | ST | 27.3±3.8 | 23.9±1.2 | 28.8±2.1 | 35.1±1.9 | 33.4±1.0 | 37.6±1.7 | **38.7±1.6** | - | 1.1% |
|   | SIFT+ST | 36.9±7.3 | 33.9±1.8 | 43.9±3.4 | 45.3±2.1 | 48.8±1.6 | 52.2±1.0 | - | **56.0±1.4** | 3.8% |
| 3 | SIFT | 42.3±5.5 | 53.9±5.6 | 50.0±5.6 | 47.2±2.6 | 52.4±1.9 | 57.1±2.3 | **57.2±2.1** | - | 0.1% |
|   | ST | 32.6±2.1 | 24.7±2.2 | 28.4±2.6 | 35.3±1.6 | 31.1±2.6 | 37.2±1.6 | **39.0±1.8** | - | 1.8% |
|   | SIFT+ST | 42.0±4.9 | 36.3±3.4 | 44.1±3.6 | 46.9±2.5 | 53.8±2.9 | 58.2±1.9 | - | **60.3±1.8** | 2.1% |
| 5 | SIFT | 46.8±4.1 | 54.9±5.2 | 53.3±5.9 | 49.0±8.1 | 54.8±7.6 | **57.4±9.0** | 57.4±8.1 | - | 0% |
|   | ST | 35.4±3.6 | 25.1±2.1 | 29.6±2.2 | 37.7±2.3 | 33.3±3.1 | 41.6±7.0 | **43.1±6.4** | - | 1.5% |
|   | SIFT+ST | 48.4±3.4 | 39.2±2.4 | 48.8±4.3 | 44.2±6.0 | 58.1±8.4 | 57.7±9.0 | - | **62.4±7.9** | 4.7% |
| 7 | SIFT | 66.5±2.7 | 71.8±3.9 | 71.9±3.8 | 62.1±2.2 | 71.6±4.5 | 72.6±4.4 | **72.8±3.5** | - | 0.2% |
|   | ST | 42.2±3.2 | 24.9±1.3 | 30.4±0.7 | **46.3±2.0** | 37.4±1.6 | 41.2±2.6 | **49.8±4.6** | - | 3.5% |
|   | SIFT+ST | 63.8±2.5 | 54.0±4.0 | 67.8±2.3 | 58.4±3.7 | 72.9±4.5 | 73.2±4.6 | - | **76.5±4.4** | 3.3% |
| 10 | SIFT | 68.7±2.7 | 73.3±3.5 | 74.0±3.8 | 66.0±2.9 | 73.6±2.6 | **74.4±2.3** | 74.7±2.5 | - | 0.3% |
|   | ST | 46.0±4.7 | 25.2±3.4 | 30.5±3.0 | 45.6±4.0 | 39.2±3.9 | 42.0±8.2 | **48.0±3.5** | - | 2.4% |
|   | SIFT+ST | 60.0±5.6 | 59.7±2.3 | 69.1±2.5 | 58.5±4.0 | 76.5±2.2 | **74.9±2.1** | - | **77.3±2.9** | 2.4% |

TABLE III
COMPARISON WITH BASELINES ON 3DA DATASET

| Source | Target | SVM_T | SGF | GFK | EDA |
|---|---|---|---|---|---|
| Amazon | Dslr | 52.9±3.1 | 44.3±4.1 | 55.9±2.5 | **62.3±2.4** |
| Amazon | Webcam | 30.5±2.8 | 39.3±1.8 | 50.9±1.9 | **55.1±2.4** |
| Webcam | Amazon | 6.90±1.8 | 19.5±1.6 | 22.1±0.6 | **23.8±2.2** |
| Webcam | Dslr | 52.9±3.1 | 47.8±2.4 | 57.0±4.4 | **60.9±2.2** |
| Dslr | Amazon | 6.90±1.8 | 10.3±1.2 | 21.1±1.0 | **24.0±2.2** |
| Dslr | Webcam | 30.5±2.8 | 36.3±2.4 | 52.9±2.2 | **55.0±2.4** |

■ **Results on 3DA Dataset**

The preliminary comparisons with several baseline methods including SVM, SGF and GFK on 3DA dataset across domains are reported in Table III. It clearly shows that the proposed EDA is much better than the competitive GFK method. Further, we conduct the comparisons with several state-of-the-art methods on the Dslr target domain (*standard setting*) in Table IV, in which the results of the compared methods can be captured in [21]. From Table IV, we observe that:

- HeMap performs the worst recognition among all methods. The reason may be that the learning process of a feature mapping matrix does not exploit the label information of target data. Thus, the learned space cannot well preserve the similar structural information of the data in both domains.
- T-SVM is slightly better than SVM_T, DAMA and ARC-t. This demonstrates that it is useful to minimize the training error of source and target data for learning without feature transformation. Notably the feature dimension of source and target data is different, so SVM_ST is not presented.
- Both HFA and SHFA are better than T-SVM for *amazon→dslr* (about 3% improvement) and *webcam→dslr* (about 2% improvement). One possible explanation is that SHFA can handle the unlabeled target data, which is not considered in other methods. Additionally, SHFA can train a better classifier by learning the transformation metric from augmented features and well exploit the source data.
- Our proposed EDA framework significantly outperforms other methods for both cases (62.3% for A→D and 62.5% for W→D). The improvement is 5.7% and 6.6% for both cases by comparing with SHFA. The improvements demonstrate that learning a classifier and a category transformation matrix simultaneously in a semi-supervised and multi-view framework can effectively promote the domain adaptability.

■ **Results on 4DA Dataset**

We then test our proposed method on the *4DA* dataset with four domains (i.e., A: *amazon*, W: *webcam*, D: *dslr*, C: *caltech*). The results of the baseline methods and state-of-the-arts that have been widely tested on this dataset are reported in Table V. For example, C→D represents that *caltech* is viewed as source domain and *dslr* is target domain. Totally, 12 cross tasks are given. From Table V, we have the following observations:

- SVM_S performs around 35% for each domain pair, except the case W→D (66.6%) and D→W (74.3%). This is due to that the shift between *dslr* and *webcam* is significantly less than the shifts between other domain pairs. Additionally, the tasks of W→D and D→W also perform the best for other approaches except the SVM_T, because SVM_T is learned using only a very limited number of target training data.
- HFA has the worst performance on the cases W→D and D→W with smaller domain shifts. The reason is that the feature augmentation in HFA may change the distribution in feature space. For the case when the domain shift is small, overfitting is easily caused in feature transformation learning, such that the performance is degraded.
- Comparatively, Symm, ARC-t, GFK and LandMark perform better than MMDT on W→D and D→W. This demonstrates that overfitting may be caused during feature transformation learning of MMDT when domain shift is too small. However, MMDT performs better in most cases with large domain shift than other methods except the proposed EDA. GFK obtains relatively good results and it may be more suitable to the case with small domain shift. Notably, the results with Dslr as source data were not reported by LandMark [55].

TABLE IV
RECOGNITION ACCURACIES (%) FOR ALL METHODS ON THE *3DA* OFFICE DATASET

| Source | Target | SVM_T | HeMap | DAMA | ARC-t | T-SVM | HFA | SHFA | EDA_SVM | Improvement |
|---|---|---|---|---|---|---|---|---|---|---|
| Amazon-800 | Dslr-600 | 52.9±3.1 | 42.8±2.4 | 53.3±2.3 | 53.1±2.4 | 53.5±2.0 | 55.4±2.9 | 56.0±2.2 | 62.3±2.4 | 5.7% |
| Webcam-800 | Dslr-600 | 52.9±3.1 | 42.2±2.6 | 53.2±3.2 | 53.0±3.2 | 53.5±2.0 | 54.3±3.6 | 55.9±3.3 | 62.5±2.2 | 6.6% |

TABLE V
RECOGNITION ACCURACIES (%) FOR ALL METHODS ON THE *4DA* EXTENDED *OFFICE* DATASET WITH LOW-LEVEL SURF FEATURE

| Method | SVM_S | SVM_T | LandMark | SGF | GFK | HFA | ARC-t | Symm | MMDT | EDA_SVM | Improvement |
|---|---|---|---|---|---|---|---|---|---|---|---|
| C→D | 35.6±0.7 | 55.8±0.9 | 57.3 | 50.2±0.8 | 57.7±1.1 | 51.9±1.1 | 50.6±0.8 | 48.6±1.1 | 56.5±0.9 | 59.0±1.2 | 1.3% |
| C→W | 30.8±1.1 | 60.3±1.0 | 49.5 | 54.2±0.9 | 63.7±0.8 | 60.5±0.9 | 50.5±1.6 | 63.8±1.1 | 67.3±0.8 | | 3.5% |
| C→A | 35.9±0.4 | 45.3±0.9 | 56.7 | 42.0±0.5 | 44.7±0.8 | 45.5±0.9 | 44.1±0.6 | 43.8±0.6 | 49.4±0.8 | 53.5±0.5 | - |
| A→C | 35.1±0.3 | 32.0±0.8 | 45.5 | 37.5±0.4 | 36.0±0.5 | 31.1±0.6 | 37.0±0.4 | 39.1±0.5 | 36.4±0.8 | 43.8±0.4 | - |
| A→W | 33.9±0.7 | 62.4±0.9 | 46.1 | 54.2±0.8 | 58.6±1.0 | 61.8±1.1 | 55.7±0.9 | 51.0±1.4 | 64.6±1.2 | 68.9±1.0 | 4.3% |
| A→D | 35.5±0.8 | 55.9±0.8 | 47.1 | 46.9±1.1 | 50.7±0.8 | 52.7±0.9 | 50.2±0.7 | 47.9±1.4 | 56.7±1.3 | 57.6±1.0 | 0.9% |
| W→C | 31.3±0.4 | 30.4±0.7 | 35.4 | 32.9±0.7 | 31.1±0.6 | 29.4±0.6 | 31.9±0.5 | 34.0±0.5 | 32.2±0.8 | 38.6±0.5 | 3.2% |
| W→A | 31.3±0.4 | 45.6±0.7 | 40.2 | 43.4±0.7 | 44.1±0.4 | 45.9±0.7 | 43.4±0.5 | 43.7±0.7 | 47.7±0.9 | 52.4±0.9 | 4.7% |
| W→D | 66.6±0.7 | 55.1±0.8 | 75.2 | 78.6±0.4 | 70.5±0.7 | 51.7±1.0 | 71.3±0.8 | 69.8±1.0 | 67.0±1.1 | 73.8±0.8 | - |
| D→C | 31.4±0.3 | 31.7±0.6 | - | 32.9±0.4 | 32.9±0.5 | 31.0±0.5 | 33.5±0.4 | 34.9±0.4 | 34.1±0.8 | 38.0±0.4 | 3.1% |
| D→A | 34.0±0.3 | 45.7±0.9 | - | 42.9±0.7 | 45.7±0.6 | 45.8±0.9 | 42.5±0.5 | 42.7±0.5 | 46.9±1.0 | 50.4±0.8 | 3.5% |
| D→W | 74.3±0.5 | 62.1±0.8 | - | 78.6±0.4 | 76.5±0.5 | 62.1±0.7 | 78.3±0.5 | 78.4±0.9 | 74.1±0.8 | 84.1±0.6 | 5.5% |
| Average | 40.0±0.6 | 48.5±0.8 | 50.3 | 49.7±0.7 | 51.0±0.7 | 47.4±0.8 | 49.5±0.6 | 48.7±0.9 | 52.5±1.0 | 57.3±0.8 | 4.8% |

TABLE VI
RECOGNITION ACCURACIES (%) ON THE *4DA* EXTENDED *OFFICE* DATASET WITH DEEP CNN-FEATURE

| Method | Layer | C→D | C→W | C→A | A→C | A→W | A→D | W→C | W→A | W→D | D→C | D→A | D→W |
|---|---|---|---|---|---|---|---|---|---|---|---|---|---|
| SVM_S | $f_6$ | 76.6±2.2 | 67.5±1.6 | 85.8±0.4 | 79.3±0.3 | 70.5±0.9 | 80.8±0.8 | 59.5±0.9 | 66.8±1.0 | 96.1±0.4 | 67.3±1.2 | 77.0±1.0 | 90.5±0.6 |
| | $f_7$ | 77.6±1.1 | 67.8±1.8 | 86.5±0.5 | 79.3±0.3 | 71.6±0.6 | 81.3±0.7 | 68.1±0.6 | 73.4±0.7 | 96.2±0.6 | 74.3±0.6 | 81.8±0.5 | 95.1±0.8 |
| SVM_T | $f_6$ | 82.0±2.8 | 73.3±3.3 | 77.5±3.5 | 55.4±2.8 | 74.2±3.5 | 77.2±4.2 | 44.0±3.9 | 75.3±3.4 | 80.2±2.6 | 55.5±2.6 | 73.4±2.8 | 67.1±3.0 |
| | $f_7$ | 85.7±2.5 | 80.0±2.1 | 83.9±2.2 | 62.4±2.8 | 79.5±2.5 | 85.8±2.7 | 57.0±3.5 | 85.5±1.5 | 83.3±2.4 | 61.2±2.6 | 82.6±2.6 | 72.4±2.9 |
| SGF | $f_6$ | 93.1±1.2 | 89.4±0.9 | 88.5±0.4 | 77.1±0.8 | 87.2±0.9 | 90.5±0.8 | 74.1±0.8 | 87.2±0.5 | 97.7±0.4 | 75.9±1.0 | 88.0±0.8 | 96.8±0.4 |
| | $f_7$ | 92.4±1.1 | 87.8±0.8 | 89.3±0.4 | 77.4±0.7 | 88.1±0.8 | 92.0±1.3 | 76.8±0.7 | 86.8±0.7 | 97.6±0.5 | 78.2±0.7 | 88.0±0.5 | 95.7±0.8 |
| GFK | $f_6$ | 92.0±1.2 | 87.7±0.8 | 87.5±0.3 | 78.9±1.1 | 89.5±0.8 | 92.6±0.7 | 77.5±0.8 | 86.2±0.8 | 97.8±0.5 | 78.8±0.8 | 88.9±0.3 | 97.0±0.8 |
| | $f_7$ | 91.9±0.8 | 86.4±0.7 | 88.4±0.4 | 79.1±0.7 | 88.6±0.8 | 94.3±0.7 | 76.1±0.7 | 85.6±0.5 | 98.5±0.3 | 77.5±0.8 | 90.1±0.4 | 96.5±0.3 |
| EDA_SVM | $f_6$ | 93.9±0.6 | 92.2±0.7 | 91.0±0.1 | 84.9±0.3 | 91.6±0.6 | 95.3±0.5 | 82.6±0.3 | 90.6±0.3 | 99.2±0.2 | 84.4±0.2 | 91.7±0.2 | 98.6±0.3 |
| | $f_7$ | 93.6±0.6 | 91.8±0.6 | 91.0±0.1 | 84.9±0.3 | 92.2±0.6 | 95.0±0.3 | 83.0±0.4 | 91.2±0.2 | 99.1±0.2 | 84.9±0.3 | 91.7±0.3 | 98.0±0.3 |
| MvEDA | $f_{67}$ | 94.1±0.6 | 92.6±0.6 | 91.6±0.1 | 85.7±0.3 | 92.4±0.6 | 95.7±0.4 | 84.0±0.4 | 91.6±0.2 | 99.3±0.1 | 85.8±0.2 | 92.3±0.2 | 98.7±0.2 |

- The results of both tasks C→W and A→W for all methods except SVM_S are ranking the second compared with that of W→D and D→W in all tasks. Interestingly, C→D and A→D are ranking the third. This demonstrates that *webcam* and *dslr* are more suitable to be target domains, while *amazon* and *caltech* are more suitable to be source domains.

- The proposed EDA outperforms other methods for most tasks. The highest improvement is 5.5% for D→W, and the average improvement is 4.8%. Note that the improvement is computed between the best and the second best results, highlighted in Table V. The highest average accuracy of EDA is 57.3%, followed by MMDT (52.5%). The significant improvements indicate that the proposed EDA is well suited to handle small or large domain shifts for object recognition.

- It's worth noting that the deep features (DeCAF) of 4DA office dataset based on CNN [52], [53] have been explored by using several state-of-the-art domain adaptation methods. The results are reported in Table VI, from which we can clearly observe that the proposed EDA outperforms other methods. By jointly learning the deep features of the 6th and 7th fully-connected layers together (*i.e.*, $f_{67}$) using the proposed MvEDA, the performance is further improved.

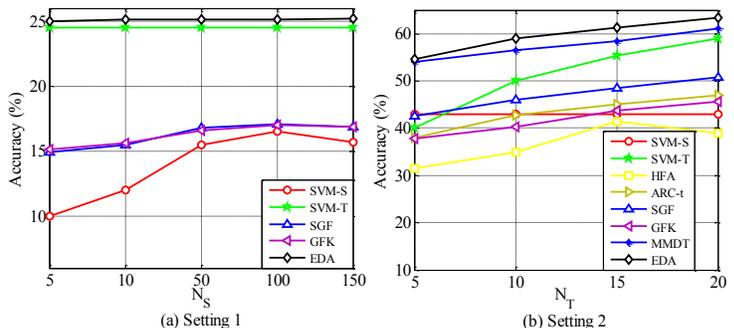

Fig. 4. Recognition accuracy on Bing-Caltech data with two settings

**Results on Bing-Caltech Dataset**

By following the two settings (Setting 1 and 2) [38], the testing accuracies are shown in Fig.4 (a) and (b), from which we can observe that the proposed EDA method still performs the best by comparing with other state-of-the-art domain adaptation methods. For setting 1, the proposed EDA has an improvement of 0.6% than SVM-T, but much higher than SGF and GFK. For setting 2, the proposed EDA is 2.5% higher than MMDT. By comparing with several popular domain adaptation

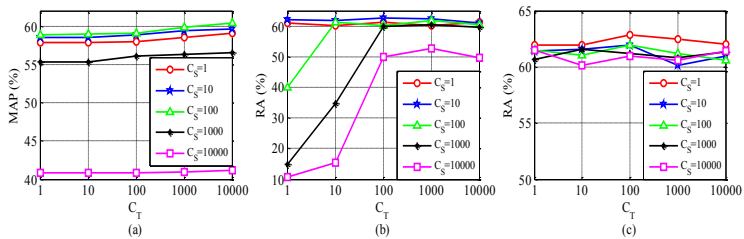

Fig. 5. Performance variations of our EDA using different parameters $C_S$ and $C_T$. (a) MAP on SIFT+ST features of YouTube & Consumer Videos data with 3 labeled target training samples per event. (b) Recognition accuracy (RA) tested on 3DA Office data for task *amazon→dslr*. (c) RA for task *webcam→dslr*.

### TABLE VII
### EMPIRICAL PARAMETER SELECTION

| Parameter | $C_S$ | $C_T$ | $\gamma$ | $\lambda$ | $\tau$ | $L$ |
|-----------|-------|-------|----------|-----------|--------|-----|
| Value | 1~10 | ≈100$C_S$ | ≈$C_S$ | ≈$C_S$ | ≈1000$C_S$ | ≈$D\log N$ |

methods on large-scale data, the performance of our EDA is proven to be comparable to state-of-the-art methods. The generalization and robustness of EDA is well shown.

### E. Parameter Sensitivity and Selection

To evaluate the performance variations of our EDA with the two free parameters $C_S$ and $C_T$, we conduct the experiments on YouTube & Consumer Videos data and Office data (two tasks: *amazon→dslr* and *webcam→dslr*). The two parameters are tuned from the set {1, 10, 100, 1000, 10000} in experiment. Specifically, we alternatively free one parameter by frozen the other one. The results for different tasks are shown in Fig. 5, from which we can observe the performance variation of our EDA with respect to the parameters. We find that the performance drops dramatically when $C_S$ is set to be a large value (*e.g.*, 1000 and 10000). From Fig.5(a) and (b), we easily obtain that $C_S \leq 100$ and $C_T \geq 100$ can be better choices for the optimal parameter selection. To this end, the empirical selection of the parameters is summarized in Table VII.

### F. Time Analysis

We present the time analysis on the first two datasets, respectively. The training time in seconds for the methods (baselines and the state-of-the-arts) is reported in Table VIII and Table IX, respectively. We observe that the proposed EDA shows very competitive time cost compared with others.

### G. Convergence

In Theorem 1 and 2, we have theoretically proved that EDA with single or multiple views is jointly convex *w.r.t.* $\boldsymbol{\beta}$, $\boldsymbol{\Theta}$, and $\alpha_v$. We have shown the convergence of the objective function and the variation $\|\boldsymbol{\beta}^t - \boldsymbol{\beta}^{t-1}\|_F$ of the classifier $\boldsymbol{\beta}$ in Fig.6, in which three datasets are conducted based on the proposed EDA. Note that Fig.6(a, d) is obtained with 3 labeled training samples per category from target domain based on SIFT+ST features by using the proposed EDA method. Fig.6(b, e) is obtained by referring *webcam* as source domain and *dslr* as target domain. Fig.6(c, f) is an example that refers *amazon* as source domain and *dslr* as target domain, *i.e.*, A→D. As can be seen from the convergence behavior shown in Fig.6(a)~(c), we find that EDA can converge to a stable point after several iterations and the convergence is demonstrated.

### TABLE VIII
### TRAINING TIME (S) ON THE YOUTUBE & CONSUMER VIDEOS DATA

|  | SVM T | SVM ST | FR | MKL | DTMKL | AMKL | EDA |
|--|-------|--------|-----|------|-------|------|-----|
| Time | 18.03 | 34.4 | 70.3 | 98.1 | 179.39 | 194.3 | 36.4 |

### TABLE IX
### TRAINING TIME (S) ON THE *3DA OFFICE* DATA

|  | SVM T | HeMap | DAMA | ARC-t | TSVM | HFA | SHFA | EDA |
|--|-------|-------|------|-------|------|-----|------|-----|
| Time | 0.06 | 1.92 | 48.86 | 11.88 | 124.2 | 19.0 | 216.7 | 5.4 |

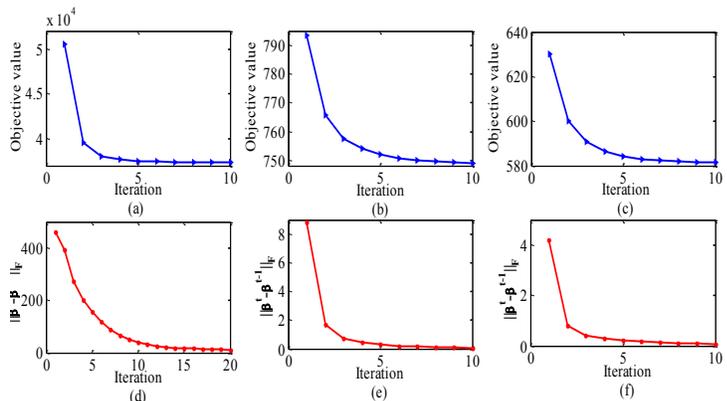

Fig. 6. Convergence of EDA for three datasets: (a, d) Consumer & YouTube Videos; (b, e) *3DA* Office dataset: *webcam→dslr*; (c, f) *4DA* Extended Office dataset: *amazon→dslr*

### VI. CONCLUSION AND FUTURE WORK

A new cross-domain learning method called Extreme Learning Machine based Domain Adaptation (EDA) is proposed in this paper. Specifically, we simultaneously learn a network classifier and a category transformation by using labeled source data, a limited number of target data and unlabeled target data. Additionally, we extend EDA to a joint learning framework of multiple views for structural information sharing of multiple local features with different feature representations. Promising results and theoretical proofs guarantee and demonstrate the efficacy of EDA for visual recognition.

Cross-domain learning is still a challenging research topic of computer vision. The problem addressed in this paper supposes that the categories from both domains are weakly similar, which may not hold in real-world scenarios. The problems provided that new classes are generated in target are noticed.

### APPENDIX

*Proofs* of *claims* 1~5 are referred as **Supplementary Material**.

### ACKNOWLEDGMENTS

We are grateful to the AE and anonymous reviewers for their valuable comments on our work, which have greatly improved the quality of our paper.